\documentclass[10pt,twocolumn,letterpaper]{article}

\usepackage{cvpr}              

\usepackage{graphicx}
\usepackage{amsmath}
\usepackage{amssymb}
\usepackage{booktabs}

\usepackage[pagebackref,breaklinks,colorlinks]{hyperref}

\usepackage[capitalize]{cleveref}
\crefname{section}{Sec.}{Secs.}
\Crefname{section}{Section}{Sections}
\Crefname{table}{Table}{Tables}
\crefname{table}{Tab.}{Tabs.}

\usepackage{titletoc}

\usepackage{amsmath,amsfonts,bm}

\def\eqref#1{equation~\ref{#1}}

\def\1{\bm{1}}

\def\rs{{\textnormal{s}}}

\def\rw{{\textnormal{w}}}

\def\rz{{\textnormal{z}}}

\def\vk{{\bm{k}}}

\def\vs{{\bm{s}}}

\def\vz{{\bm{z}}}

\def\mI{{\bm{I}}}

\def\mK{{\bm{K}}}

\DeclareMathAlphabet{\mathsfit}{\encodingdefault}{\sfdefault}{m}{sl}
\SetMathAlphabet{\mathsfit}{bold}{\encodingdefault}{\sfdefault}{bx}{n}

\newcommand{\pz}{p_{\rm{z}}}

\DeclareMathOperator*{\argmax}{arg\,max}

\usepackage{url}
\usepackage[utf8]{inputenc} 
\usepackage[T1]{fontenc}    
\usepackage{xcolor}         
\definecolor{mydarkblue}{rgb}{0,0.08,0.45}
\usepackage{hyperref}       
\hypersetup{ %
    pdftitle={},
    pdfauthor={},
    pdfsubject={},
    pdfkeywords={},
    pdfborder=0 0 0,
    pdfpagemode=UseNone,
    colorlinks=true,
    linkcolor=black,
    citecolor=mydarkblue,
    filecolor=mydarkblue,
    urlcolor=mydarkblue,
    pdfview=FitH}
\usepackage{url}            
\usepackage{booktabs}       
\usepackage{amsfonts}       
\usepackage{nicefrac}       
\usepackage{microtype}      

\usepackage{multirow}
\usepackage{bigstrut}
\usepackage{graphicx}
\usepackage{amsmath}
\usepackage{amssymb}
\usepackage{algorithm}
\usepackage{algorithmicx}
\usepackage{algpseudocode}
\usepackage{tabu}
\usepackage{siunitx}
\usepackage{adjustbox}
\usepackage{siunitx}
\usepackage{makecell}
\usepackage{colortbl}
\usepackage{color}
\usepackage{multirow}
\usepackage{blindtext}
\usepackage{comment}
\usepackage{bm}
\usepackage{bbm}
\usepackage{booktabs}
\usepackage{wrapfig}
\usepackage{array} 
\definecolor{Gray}{gray}{0.95}
\definecolor{Cyan}{rgb}{0.88,1,1}
\definecolor{LightCyan}{rgb}{0.92,1,1}
\definecolor{DarkCyan}{rgb}{0.82,1,1}
\definecolor{DarkRed}{rgb}{1,0.82,1}
\definecolor{LightRed}{rgb}{1, 0.92,1}
\usepackage{float}
\usepackage{wasysym}
\usepackage{pifont}

\newcommand{\ourmeos}{\textbf{\texttt{BOsampler}} }

\usepackage{wrapfig}

\def\cvprPaperID{414} 
\def\confName{CVPR}
\def\confYear{2023}

\begin{document}


\newcommand{\ourtitle}{Unsupervised Sampling Promoting for Stochastic Human Trajectory Prediction}
\title{\ourtitle}


\author{
  Guangyi Chen\thanks{Authors contributed equally and are listed alphabetically by first name.}\textsuperscript{\space\space ,\ding{169},\ding{168}}, Zhenhao Chen\textsuperscript{*,\ding{169}}, Shunxing Fan\textsuperscript{\ding{169}}, Kun Zhang \textsuperscript{\ding{169},\ding{168}}\\
  \textsuperscript{\ding{169}}Mohamed bin Zayed University of Artificial Intelligence \\ \textsuperscript{\ding{168}}Carnegie Mellon University
}

\maketitle

\begin{abstract}
The indeterminate nature of human motion requires trajectory prediction systems to use a probabilistic model to formulate the multi-modality phenomenon and infer a finite set of future trajectories. However, the inference processes of most existing methods rely on Monte Carlo random sampling, which is 
insufficient to cover the realistic paths with finite samples, due to the long tail effect of the predicted distribution.
To promote the sampling process of stochastic prediction, we propose a novel method, called \ourmeos, to adaptively mine potential paths with Bayesian optimization in an unsupervised manner, as a sequential design strategy in which new prediction is dependent on the previously drawn samples.
Specifically, we model the trajectory sampling as a Gaussian process and construct an acquisition function to measure the potential sampling value. This acquisition function applies the original distribution as prior and encourages exploring paths in the long-tail region. This sampling method can be integrated with existing stochastic predictive models without retraining. Experimental results on various baseline methods demonstrate the effectiveness of our method. The source code is released in this \href{https://github.com/viewsetting/Unsupervised_sampling_promoting}{link}.
\end{abstract}

\section{Introduction}
\label{sec:intro}

Humans usually behave indeterminately due to intrinsic intention changes or external surrounding influences. It requires human trajectory forecasting systems to formulate humans' multimodality nature and infer not a single future state but the full range of plausible ones~\cite{mid,ynet}.

Facing this challenge, many  prior methods formulate stochastic human trajectory prediction as a generative problem, in which a latent random variable is used to represent multimodality. A typical category of methods~\cite{socialgan,Sun_2020_CVPR,zhao2019multi,dendorfer2021mg} is based on generative adversarial networks (GANs), which generate possible future trajectories by a noise in the multi-modal distribution. Another category exploits the variational auto-encoder (VAE)~\cite{trajectron++,ivanovic2019trajectron,lee2017desire,tang2019multiple,Liu_2021_ICCV} that uses the observed history trajectories as a condition to learn the latent variable. Beyond these two mainstream categories, other generative models are also employed for trajectory prediction, such as diffusion model~\cite{mid}, normalized flow~\cite{rhinehart2018r2p2}, and even simple Gaussian model~\cite{stgcnn,sgcn}.


\begin{figure}[t]
    \centering
    \includegraphics[width=\linewidth,trim=0 80 280 0,clip]{ 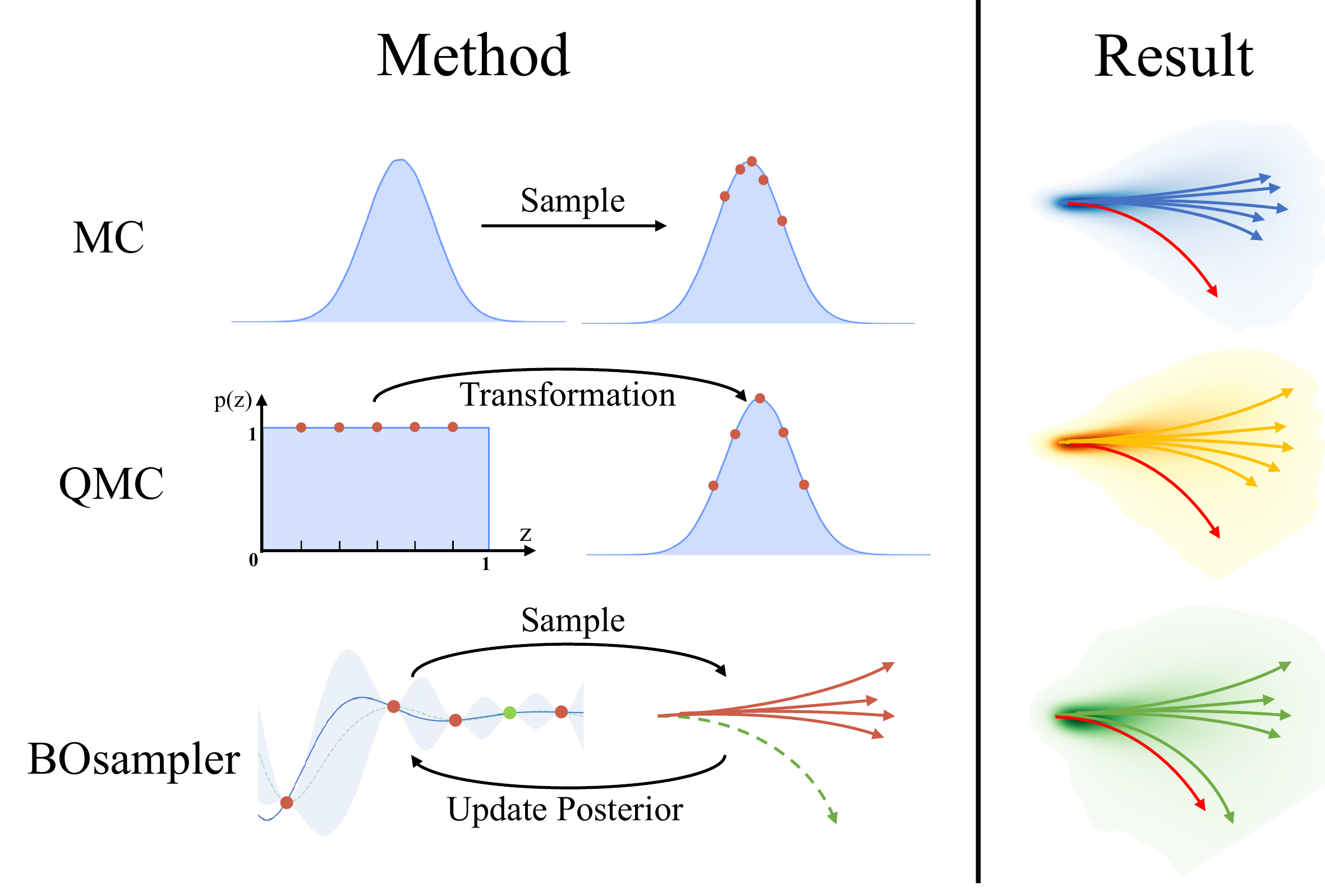}
    \caption{The comparison of different sampling methods. Monte Carlo (MC) sampling generates trajectories by directly sampling from a prior distribution of latent variable $z$. Quasi-Monte Carlo (QMC) sampling uses a transformation from low-discrepancy sequences to the prior distribution \cite{niederreiter1992random} to sample more uniformly than MC. Different from MC and QMC, \ourmeos formulates the sampling process as a Gaussian Process and calculate the Gaussian posterior with existing samples to sample the next one, where sampling and posterior updating are iterative.
    }
    \label{fig: top}
    \vspace{-0.3cm}
\end{figure}

Instead of a single prediction, the inference process of these stochastic prediction methods produces a finite set of plausible future trajectories by Monte Carlo (MC) random sampling. However, the distributions are always uneven and biased, where the common choices like ``go straight'' are in high probability. In contrast, many other choices such as ``turn left/right'' and ``U-turn'' are in low probability.
Due to the long tail effect of predicted distribution, finite samples are insufficient to cover the realistic paths. 
For example, as shown in Figure~\ref{fig: top}, MC sampling tends to generate redundant trajectories with high probability but ignores the potential low-probability choice. 
To solve this problem, some methods~\cite{ma2021likelihood,npsn} trained the model using an objective term to increase the diversity of samples, e.g., maximizing the distance among the predicted samples. Though improving the sampling diversity, these methods need to re-train the model by adding the loss term. It is timely-cost and may fail when only the model is given (the source data is inaccessible).

In this paper, we propose an unsupervised method to promote the sampling process of stochastic prediction without accessing the source data. It is named \ourmeos, which refines the sampling for more exploration via Bayesian optimization (BO).  
Specifically, we first formulate the sampling process as a Gaussian Process (GP), where the posterior is conditioned by previous sampling trajectories. Then, we define an acquisition function to measure the value of potential samples, where the samples fitting the trained distribution well or away from existing samplings obtain high values. By this acquisition function, we can encourage the model to explore paths in the long-tail region and achieve a trade-off between accuracy and diversity. 
As shown in Figure~\ref{fig: top}, we compare \ourmeos with MC and another sampling method  QMC~\cite{npsn}, which first generates a set of latent variables from a uniform space and then transfers it to prior distribution for trajectory sampling. 
Compared with them, \ourmeos can adaptively update the Gaussian
posterior based on existing samples, which is more flexible.
We highlight that \ourmeos serves as a plug-and-play module that could be integrated with existing multi-modal stochastic predictive models to promote the sampling process without retraining.
In the experiments, we apply the \ourmeos on many popular baseline methods, including Social GAN~\cite{socialgan}, PECNet~\cite{pecnet}, Trajectron++~\cite{trajectron++}, and Social-STGCNN~\cite{stgcnn}, and evaluate them on the ETH-UCY datasets.
The main contributions of this paper are summarized as follows:
\begin{itemize}
\setlength{\itemsep}{1pt}
\setlength{\parsep}{1pt}
\setlength{\parskip}{1pt}
    \item We present an unsupervised sampling prompting method for stochastic trajectory prediction, which mines potential plausible paths with Bayesian optimization adaptively and sequentially.
    \item The proposed method can be integrated with
existing stochastic predictors without retraining. 
    \item We evaluate the method with multiple baseline methods and show significant improvements.
\end{itemize}


\section{Related Work}

\textbf{Trajectory Prediction with Social Interactions.}
The goal of human trajectory forecasting is to infer plausible future positions with the observed human paths. In addition to the destination, the pedestrian's motion state is also influenced by the interactions with other agents, such as other pedestrians and the environment. Social-LSTM~\cite{alahi2016social} apply a social pooling layer to merge the social interactions from the neighborhoods. To highlight the valuable clues from complex interaction information, the attention model is applied to mine the key neighbourhoods~\cite{fernando2018soft+,vemula2018social,xu2022remember,zheng2021unlimited,adeli2021tripod}.
Besides, for the great representational ability of complex relations, some methods apply graph model to social interaction~\cite{huang2019stgat,Sun_2020_CVPR2,chen2021human,xu2022groupnet,xu2022adaptive,bae2022learning}. 
To better model social interactions and temporal dependencies, different model architectures are proposed for trajectory prediction, such as RNN/LSTM~\cite{zhang2019sr}, CNN~\cite{nikhil2018convolutional,stgcnn}, and Transformer~\cite{li2022graph,yu2020spatio,tsao2022social,yuan2021agentformer}.
Beyond human-human interactions, human-environment interaction is also critical to analyze human motion. To incorporate the environment knowledge, some methods encode the scene image or traffic map with the convolution neural network
~\cite{wong2022view,sun2022m2i,sun2022human,meng2022forecasting,zhong2022aware,xu2022pretram}.

\textbf{Stochastic Trajectory Prediction.} 
The above deterministic trajectory prediction methods only generate one possible prediction, ignoring human motion's multimodal nature. To address this problem, stochastic prediction methods are proposed to represent the multimodality by the generative model. Social GAN~\cite{socialgan} first introduces the Generative adversarial networks (GANs) to model the indeterminacy and predict socially plausible futures. In the following, some GAN-based methods are proposed to integrate more clues~\cite{sadeghian2019sophie,dendorfer2021mg} or design more efficient models~\cite{kosaraju2019social,Fang_2020_CVPR,Sun_2020_CVPR}.
Another kind of methods~\cite{ivanovic2019trajectron,Chen_2021_ICCV,lee2022muse,chen2022scept,xu2022socialvae,halawa2022action,yue2022human} formulates the trajectory prediction as CVAE~\cite{sohn2015learning}, which applies observed trajectory as condition and learn a latent random variable to model multimodality. Besides, some methods explicitly use the endpoint
~\cite{pecnet,zhao2020tnt,ynet,zhao2021you,gu2021densetnt,girase2021loki} to model the possible destinations or learn the grid-based location encoder~\cite{liang2020garden,deo2020trajectory,guo2022end} generate acceptable paths. Another 
Recently, Gu \emph{et al.}~\cite{mid} proposes to use the denoise diffusion probability model(DDPM) to discard the indeterminacy gradually to obtain the desired trajectory region.
Beyond learning a better probability distribution of human motion, some methods~\cite{ma2021likelihood,npsn} focus on learning the sampling network to generate more diverse trajectories. However, these methods need to retrain the model, which is timely-cost and can only work when source data is given.

\textbf{Bayesian Optimization.} 
The key idea of Bayesian optimization (BO)~\cite{shahriari2015taking}
is to drive optimization decisions with an adaptive model.
Fundamentally, it is a sequential model to find the global optimization result of an unknown objective function. Specifically, it initializes a prior belief for the objective function and then sequentially updates this model with the data selected by Bayesian posterior. BO has emerged as an excellent tool in a wide range of fields, such as hyper-parameters tuning~\cite{snoek2012practical}, automatic machine learning~\cite{white2021bananas,kandasamy2018neural}, and reinforcement learning~\cite{brochu2010tutorial}. Here, we introduce BO to prompt the sampling process of stochastic trajectory forecasting models. We formulate the sampling process as a sequential Gaussian process and define an acquisition function to measure the value of potential samples. With BO, we can encourage the model to explore paths in the long-tail regions.


\section{Method}
In this section, we will introduce our unsupervised sampling promoting method, \ourmeos, which is motivated by Bayesian optimization to sequentially update the sampling model given previous samples.
First, we formulate the sampling process as a Gaussian process where new samples are conditioned on previous ones. Then we show how to adaptively mine the valuable trajectories by Bayesian Optimization. Finally, we provide a detailed optimization algorithm of our method.
\subsection{Problem Definition} 
Given observed trajectories for $L$ pedestrians with time steps t=1,...,$T_{obs}$ in a scene as $X_l^{1:T_{obs}}=\{X_l^t|t \in [1,...,T_{obs}]\}$ for $\forall l \in [1,...,L]$, 
the trajectory predictor will generate N possible future trajectories $\hat{Y}_l^{1:T_{pred}}=\{\hat{Y}_{l,n}^t | t\in [1,...,T_{pred}],n\in [1,...,N]\}$ for each pedestrian, where $X_l^t $ and $Y_l^t $ are both 2D locations. 
For the sake of simplicity, we remove the pedestrian index $l$ and time sequences $1:T_{obs} $ and $1:T_{pred}$ without special clarification, e.g., using $X$ and $\hat{Y}_n$ to respectively represent the observed trajectory and one of the generated future paths. Then the trajectory prediction system can be formulated as:
\begin{equation}
\setlength{\abovedisplayskip}{1pt}
\setlength{\belowdisplayskip}{1pt}
\begin{aligned}
\label{eq: generator}
\hat{Y} := G_{\theta}(X,\rz),
\end{aligned}
\end{equation}
where $G_{\theta} $ denotes a predictor with learned parameters $\theta$ , and $ \rz $ is a latent variable with distribution $\pz$ to model the multimodality of human motion. For a GAN-based model, $\pz$ is a multivariate normal distribution, and for a CVAE-based model, $\pz$ is a latent distribution. In the inference stage, we will sample a sequence of values of latent variable $\{\rz_n\}_{n=1}^{N} \in \pz$ to generate a finite set of future trajectories as $\hat{Y}_n := G_{\theta}(X,\rz_n) $.


\begin{figure*}
    \centering
    \includegraphics[width=\linewidth,trim=0 375 0 0,clip]{ 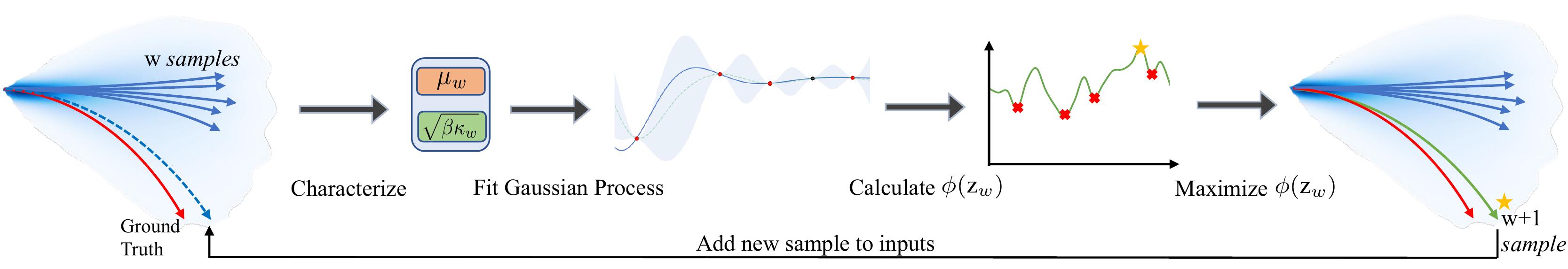}
    \caption{An illustration of how \ourmeos generate new samples in an iterative manner. Given w existing samples, we first characterize two functions: $\mu_w$ and $\sqrt{\beta \kappa_w}$. Then, we use these two functions to fit a Gaussian Process and calculate the posterior distribution. Next, we use the posterior distribution to calculate the acquisition function $\phi(\text{z}_w)$. Then, the next sample is generated by maximizing $\phi(\text{z}_w)$.
    Finally, we add this new sample to inputs and can use it to update the distribution.}
    \label{fig:framework}
    \vspace{-0.1cm}
\end{figure*}



\subsection{BOsampler}

Conventional stochastic trajectory prediction methods sample latent variables in a Monte Carlo manner based on learned distribution. Despite learning well, the distributions are always uneven and biased, where the common choices like ``go straight'' are in high probability and other choices such as ``U-turn'' are in low probability. Due to this long-tail characteristic of distribution, finite trajectories with overlapped high-probability paths and less low-probability paths cannot cover the realistic distribution. Though low-probability situations are the minority in the real world, they may raise potential serious safety problems, which are important for the applications such as auto-driving. 

To solve this problem, we propose to select valuable samples with Bayesian optimization. The optimization objective can be formulated as:
\begin{equation}
\setlength{\abovedisplayskip}{2pt}
\setlength{\belowdisplayskip}{2pt}
\begin{aligned}
\label{eq: objective}
\rz^* = \argmax\limits_{\rz\in \mathcal{Z}}\sum_{l=1}^{L}D(G_{\theta}(X_l,\rz)),
\end{aligned}
\end{equation}
where $G_{\theta}(X_l,\rz)$ denotes the generated trajectory and $D$ is a metric to evaluate the quality of the sampling, e.g., the average distance error (ADE) of the trajectory. Given the sampling space, $\rz \in \mathcal{Z}$, the goal of \ourmeos is to find a $\rz^*$ to achieve the best score of the evaluation metric using a finite number of samples. For simplicity, we define the above objective function as $ f(\rz)=\sum_{l=1}^{L}D(G_{\theta}(X_l,\rz)) $.

\subsubsection{Gaussian Process}
To optimize the $f(\rz)$, we formulate the sampling as a sequential Gaussian process defined on the domain $\mathcal{Z}$, which is characterized by a mean function $\mu(\rz): \mathcal{Z} \rightarrow \mathbb{R}$ and a covariance  function $\kappa(\rz): \mathcal{Z}^2 \rightarrow \mathbb{R}$ (defined by the kernel). This Gaussian process can serve as a 
probabilistic surrogate of the objective function as:
\begin{equation}
\setlength{\abovedisplayskip}{2pt}
\setlength{\belowdisplayskip}{2pt}
\begin{aligned}
\label{eq: gp}
f(\rz) \sim \mathcal{G} \mathcal{P}\left(\mu(\rz), \kappa(\mathbf{z}, \mathbf{z}^{\prime})\right),
\end{aligned}
\end{equation}
where 
\begin{equation}
\setlength{\abovedisplayskip}{2pt}
\setlength{\belowdisplayskip}{2pt}
\begin{aligned}
\label{eq: mu_cov}
\mu(\rz) &=\mathbb{E}[f(\rz)] \\
\kappa(\mathbf{z}, \mathbf{z}^{\prime}) &=\mathbb{E}\left[(f(\rz)-\mu(\rz))^T(f(\rz^\prime)-\mu(\rz^{\prime}))\right].
\end{aligned}
\end{equation}
Given the previous $w-1$ generated paths and corresponding evaluation scores $\Omega_{w-1}= \{(\rz_i, \rs_i)\}_{i=1}^{w-1}$ 
, where $\rz_i \in \mathcal{Z}$ is the sample
and $\rs_i = f(\rz_i)+\epsilon_i \in \mathbb{R}$ is the real evaluation score with possible measure noise $\epsilon \sim \mathcal{N}(0,\delta^2) $, we want to calculate the distribution of the next generated sample and score $(\rz_{w}, f(\rz_{w})) $.
Define that the vectors $\vs=[\rs_1,\rs_2,...,\rs_{w-1}]^T \in \mathbb{R}^{w-1}$, $\vz=[\rz_1,\rz_2,...,\rz_{w-1}]^T \in \mathbb{R}^{w-1}$. Then Define that 
the kernel matrix $\mK \in \mathbb{R}^{(w-1)\times (w-1)}$ with $\mK_{i,j} = \kappa(\rz_i,\rz_j) $, and $\vk,\vk^{\prime} \in \mathbb{R}^{w-1} $ are two vectors from $\mK$ as $\vk_i = \kappa(\rz,\rz_i)$ and $\vk^{\prime}_i = \kappa(\rz^{\prime},\rz_i)$.
The joint distribution of previous scores $\vs $ and the next score $\rs_{w}$ can be formulated as:
\begin{equation}
\setlength{\abovedisplayskip}{2pt}
\setlength{\belowdisplayskip}{2pt}
\begin{aligned}
\label{eq: joint}
\left[\begin{array}{l}
\vs \\
f(\rz_w)
\end{array}\right] \sim \mathcal{N}\left(\left[\begin{array}{l}
\mu(\vz) \\
\mu(\rz_{w})
\end{array}\right],\left[\begin{array}{ll}
\mK & \vk^{\prime} \\
\vk^T & \kappa(\rz_w,\rz_w)
\end{array}\right]\right).
\end{aligned}
\end{equation}
Here, the posterior distribution $f(\rz_w)|\Omega_{w-1} \sim \mathcal{N}(\mu_w(\rz_w),\kappa_w(\rz_w) ) $ is still a Gaussian distribution by utilizing the properties of Gaussian process.
The mean and covariance functions of this posterior distribution can be formulated as:
\begin{equation}
\setlength{\abovedisplayskip}{2pt}
\setlength{\belowdisplayskip}{2pt}
\begin{aligned}
\label{eq: mu_cov_post}
\mu_w(\rz_w) &=\mu(\rz_w)+\vk^T(\mK+\sigma^2\mI)^{-1}(\vs-\mu(\vz)) \\
\kappa_w(\rz_w, \rz_w) &=
\kappa(\rz_w, \rz_w)- \vk^T(\mK+\sigma^2\mI)^{-1}\vk^{\prime}.
\end{aligned}
\end{equation}
This closed-form solution of the posterior process indicates that we can easily update the probabilistic model of $f(\rz)$ with new sample $\rz_w$. 
As shown in Figure~\ref{fig:framework}, we can iteratively use the posterior distribution to select new samples and use new samples to update the distribution. 
Specifically, given the $w-1$ sampled trajectories and the corresponding latent variables, we first obtain a database $\Omega_{w-1}= \{(\rz_i, \rs_i)\}_{i=1}^{w-1}$. 
Then we can calculate the posterior distribution of the possible evaluation score $f(\rz_w)$, and use this posterior distribution to select the next sample $\rz_w$. 
Then we add this sample to the database to obtain $\Omega_{w}$ and further select $\rz_{w+1}$.

\subsubsection{Acquisition Function}
To select the next sample, we apply this posterior distribution to define an acquisition function $\phi(\rz)$ to measure the value of each sample. On the one hand, the good samples deserve a high evaluation score $f(\rz)$. On the other hand, we encourage the model to explore the regions never touched before.
To achieve this goal, we define the acquisition function $\phi(\rz)$ as:
\begin{equation}
\setlength{\abovedisplayskip}{2pt}
\setlength{\belowdisplayskip}{2pt}
\begin{aligned}
\label{eq: acquistion}
\phi(\rz_w) = \mu_w(\rz_w) + \sqrt{\beta \kappa_w(\rz_w,\rz_w^{\prime})},
\end{aligned}
\end{equation}
where the first term denotes that we would like to select the samples with high score expectations, and the second term indicates to select the samples with more uncertainty (variance). Both two terms come from the posterior distribution.  We use a hyper-parameter $\beta$ to balance the accuracy (high score expectation) and diversity (high uncertainty). We then maximize this acquisition function as $ \rz_w^* = \argmax \phi(\rz_w)$ to select the next samples. 

However, different from typical Bayesian Optimization, our task finds the score function inaccessible since we cannot obtain the ground-truth trajectory during sampling. To solve this problem, we propose a pseudo-score evaluation function to approximate the ground-truth function. 
Specifically, we assume only slight bias exists between the training and testing environment, and the same for using the most likely predicted trajectory and the pseudo ground truth. Taking the ADE as an example, we calculate the evaluation score as:
\begin{equation}
\setlength{\abovedisplayskip}{2pt}
\setlength{\belowdisplayskip}{2pt}
\begin{aligned}
\label{eq: pseudo score}
f(\rz) = - \sum_{l=1}^{L} D_{ade}(G_{\theta}(X,\rz),G_{\theta}(X,\tilde{\rz})),
\end{aligned}
\end{equation}
where $\tilde{\rz} = \argmax \pz$ denotes the most-likely prediction. It means that we trust the trained model without any information update. \label{sec: acq_function}

\begin{table*}[t]
\caption{Quantitative results on the exception subset with Best-of-20 strategy in ADE/FDE metric. We select the abnormal trajectories from ETH-UCY to benchmark the sampling methods for abnormal situations such as turning left/right or U-turn, which is important for safety. Gain: the average performance improvement of ADE and FDE to MC, higher is better.}
\small
\vspace{-0.3cm}
\linespread{3.0}
\renewcommand\arraystretch{1.1}
\renewcommand\tabcolsep{2pt}
\begin{center}
\newcolumntype{g}{>{\columncolor{Gray}}c}
\newcolumntype{y}{>{\columncolor{LightCyan}}c}
\newcolumntype{d}{>{\columncolor{DarkCyan}}c}
\newcolumntype{r}{>{\columncolor{LightRed}}c}
\newcolumntype{f}{>{\columncolor{DarkRed}}c}
\begin{tabular}{l| c|c g c g c g c g c g y d r f}
\hline
\hline
\multirow{2}{*}{\textbf{Baseline Model}} &\multirow{2}{*}{\textbf{Sampling}} &\multicolumn{2}{c}{\textbf{ETH}} & \multicolumn{2}{c}{\textbf{HOTEL}} &  \multicolumn{2}{c}{\textbf{UNIV}} &  \multicolumn{2}{c}{\textbf{ZARA1}} & \multicolumn{2}{c}{\textbf{ZARA2}} & \multicolumn{2}{c}{\textbf{AVG}} &
\multicolumn{2}{c}{\textbf{Gain}}\\ 
\cline{3-16}
~ & &\multicolumn{1}{c}{\textbf{ADE}} & \multicolumn{1}{c}{\textbf{FDE}} &  \multicolumn{1}{c}{\textbf{ADE}} &  \multicolumn{1}{c}{\textbf{FDE}} & 
\multicolumn{1}{c}{\textbf{ADE}} &  \multicolumn{1}{c}{\textbf{FDE}} &
\multicolumn{1}{c}{\textbf{ADE}} &  \multicolumn{1}{c}{\textbf{FDE}} &
\multicolumn{1}{c}{\textbf{ADE}} &  \multicolumn{1}{c}{\textbf{FDE}} &
\multicolumn{1}{c}{\textbf{ADE}} &  \multicolumn{1}{c}{\textbf{FDE}} &
\multicolumn{1}{c}{\textbf{ADE}} &  \multicolumn{1}{c}{\textbf{FDE}} 
\\
\hline
\multirow{4}{*}{\textbf{Social-GAN}~\cite{socialgan}}& MC & 1.52 & 2.37 & 0.61 & 1.21 & 0.91 & 1.86 & 0.78 & 1.63 & 0.90 & 1.97 & 0.94 & 1.80 & / & /\\
&QMC & 1.56 & 2.74 & 0.60 & 1.12 & 0.91 & 1.85 & 0.77 & 1.60 & 0.92 & 2.00 & 0.95 & 1.86 & -1\% & -3\% \\
 &BOSampler & 1.14 & 2.04 & 0.52 & 1.03 & 0.80  & 1.62 & 0.68 & 1.41 & 0.75 & 1.60 & 0.78 & 1.54 & 18\% & 15\%  \\
\hline
\multirow{4}{*}{\textbf{Trajectron++}~\cite{trajectron++}}& MC & 0.59& 1.41 & 0.20 & 0.44 & 0.37& 0.87&0.15 &0.35 & 0.22 & 0.47 & 0.30 & 0.71 & / & / \\
&QMC & 0.61& 1.45& 0.20& 0.43  & 0.36&0.86 & 0.15 & 0.35 &0.22 & 0.48 &0.31 & 0.71 & -1\% & -1\% \\
 &BOSampler & 0.52 & 0.95 & 0.19 & 0.39 &0.30 &0.67 &0.14 & 0.33 &0.20 & 0.45 & 0.27 & 0.56 & 11\%& 21\%  \\
\hline
\multirow{4}{*}{\textbf{PECNet}~\cite{pecnet}}& MC & 2.80 & 5.38 & 0.59 & 0.94 & 1.14 & 2.04 & 0.76 & 1.52 & 0.76 & 1.51 & 1.21 & 2.33 & / & /\\
&QMC & 2.81 & 5.35 & 0.59  & 0.98 &1.13 &2.28 & 0.68 & 1.36 & 0.78 & 1.56 & 1.20 & 2.31 & 1\% & 1\% \\
 &BOSampler & 2.11 & 3.73 & 0.46 & 0.72 & 0.97 & 1.87 &0.66 &1.27 & 0.65 & 1.18 & 0.97 & 1.75 & 19\% & 25\%\\
\hline 
\multirow{4}{*}{\textbf{Social-STGCNN}~\cite{stgcnn}}& MC & 2.18 & 4.14 & 0.30 & 0.51 & 0.57 & 1.05 & 0.56 & 1.03 & 0.50 & 0.96 & 0.82 & 1.54  & / & /\\
&QMC & 2.20 & 3.66 & 0.26 & 0.42 & 0.45 & 0.80 & 0.48 & 0.86 & 0.44 & 0.79 & 0.77 & 1.31 & 7\% & 15\%\\
 &BOSampler & 0.87 & 1.13 & 0.18 & 0.32 & 0.58 & 1.06 & 0.52 & 0.96 & 0.45 & 0.86 & 0.52 & 0.87 & 37\% & 44\%  \\
\hline
\multirow{4}{*}{\textbf{STGAT}~\cite{huang2019stgat}}& MC & 1.73 & 3.49 & 0.60 & 1.10 & 0.92 & 1.94 & 0.69 & 1.41 & 0.90 & 1.87 & 0.97 & 1.96 & / & / \\
&QMC & 1.80 & 3.61 & 0.56 & 0.98 & 0.89 & 1.87 & 0.67 & 1.33 & 0.88 & 1.85 & 0.96 & 1.93 & 1\% & 2\% \\
 &BOSampler & 0.97 & 1.57 & 0.56 & 1.01 & 0.83 & 1.74 & 0.63 & 1.23 & 0.83 & 1.71 & 0.76 & 1.45 & 21\% & 26\% \\
\hline\hline
\end{tabular}
\end{center}
\label{table:exception}
\vspace{-0.3cm}
\end{table*}

\subsection{Technical Details}
To optimize the sampling process smoothly, we apply some technical tricks for our \ourmeos. 

\textbf{Warm-up.} First, we use a warm starting to build the Gaussian process. It randomly samples $\rw$ latent variables and generates trajectories to obtain the first understanding of $f(\rz)$ as prior. This warm stage is the same as the original MC random sampling. We choose the number of warm-ups as half of the total number of sampling in our experiments. In \cref{sec: warmup}, we provide quantitative analysis about the number of warm-ups

\textbf{Acquisition Function.}
For the acquisition function, we set the latent vector $z$ to a zero vector to generate the pseudo label, and use the pseudo label to obtain the pseudo-score as \eqref{eq: pseudo score}. 
Then we tune the hyper-parameter $\beta$ of the acquisition function between $[0.1,1]$ because it's within the commonly used range in applications of Bayesian Optimization. Please refer to \cref{sec: acq_function} for details.


\textbf{Calculation. }
To make \ourmeos computes on GPU as the same as the pre-trained neural networks, we use \textsc{BoTorch}\cite{balandat2020botorch} as our base implementation. Also, we modify some parts related to the acquisition function and batch computation accordingly.  

Overall, \ourmeos is an iterative sampling method. Given a set of samples, we first build the Gaussian process and obtain the posterior distribution as \eqref{eq: mu_cov_post}. Then, we calculate the acquisition function as \eqref{eq: acquistion} and generate the new sample with the latent variable $\rz^*$ with the highest acquisition value. Finally, we add the new sample into the database and repeat this loop until we get enough samples. 

\section{Experiments}
In this section, we first quantitatively compare the performance of our \ourmeos with other sampling methods using five popular methods as baselines on full ETH-UCY dataset and the hard subset of it. Then, qualitatively, we visualize the sampled trajectories and their distribution. Finally, we provide an ablation study and parameters analysis to further investigate our method.

\subsection{Experimental Setup}

\indent\textbf{Dataset.}
We evaluated our method on one of the most widely used public human trajectory prediction benchmark dataset:  
ETH-UCY~\cite{pellegrini2010improving,lerner2007crowds}. ETH-UCY is a combination of two datasets with totally five different scenes, where the ETH dataset~\cite{pellegrini2010improving} contains two scenes, ETH and HOTEL, with 750 pedestrians, and the UCY dataset~\cite{lerner2007crowds} consists of three scenes with 786 pedestrians including UNIV, ZARA1, and ZARA2. All scenes are captured in unconstrained environments such as the road, cross-road, and almost open area. In each scene, the pedestrian trajectories are provided in a sequence of world-coordinate. 
The data split of ETH-UCY follows the protocols in Social-GAN and Trajectron++~\cite{socialgan,trajectron++}.
The trajectories are sampled at 0.4 seconds interval, where the first 3.2 seconds (8 frames) is used as observed data to predict the next 4.8 seconds (12 frames) future trajectory. 

To evaluate the performance on the uncommon trajectories (e.g. the pedestrian suddenly makes a U-turn right after the observation), we select a \textbf{exception subset} consisting of the most uncommon trajectories selected from ETH/UCY. To quantify the rate of exception, we use a linear method \cite{kalman1960new}, an off-the-shelf Kalman filter, to give a reference trajectory. Since it is a linear model, the predictions can be regarded as normal predictions. Then we calculate FDE between the ground truth and reference trajectory as a metric of deviation.
If the derivation is relatively high, it means that the pedestrian makes a sudden move or sharp turn afterward.
Finally, we select the top 4\% most deviated trajectories from each dataset of UCY/ETH as the exception subset.\label{sec:exception_dataset}

\begin{table}[t]
\caption{Quantitative results on the ETH/UCY dataset with Best-of-20 strategy in ADE/FDE metric. Lower is better. * updated version of \href{https://github.com/StanfordASL/Trajectron-plus-plus/issues/26}{Trajectron++ }}
\small
\vspace{-0.4cm}
\linespread{3.0}
\renewcommand\arraystretch{1.1}
\renewcommand\tabcolsep{3pt}
\begin{center}

\newcolumntype{g}{>{\columncolor{Gray}}c}
\newcolumntype{y}{>{\columncolor{LightCyan}}c}
\newcolumntype{d}{>{\columncolor{DarkCyan}}c}
\begin{tabular}{l| c| y d}
\hline
\hline
\multirow{2}{*}{\textbf{Baseline Model}} &\multirow{2}{*}{\textbf{Sampling}} & \multicolumn{2}{c}{\textbf{AVG}} \\ 
\cline{3-4}
~ & &
\multicolumn{1}{c}{\textbf{ADE}} &  \multicolumn{1}{c}{\textbf{FDE}} 

\\
\hline
\multirow{4}{*}{\textbf{Social-GAN}~\cite{socialgan}}& MC & 0.53&1.05\\
&QMC   & 0.53 & 1.03 \\
 &BOSampler & 0.52 & 1.01 \\
&BOSampler + QMC  & 0.52 & 1.00   \\
\hline
\multirow{4}{*}{\textbf{Trajectron++}~\cite{trajectron++}}& MC & 0.21 & 0.41  \\
&QMC & 0.21 & 0.40  \\
 &BOSampler & 0.18 & 0.36  \\
&BOSampler + QMC & 0.18 & 0.36   \\
\hline

\multirow{4}{*}{\textbf{Trajectron++}~\cite{trajectron++} \textsuperscript{*}}& MC & 0.28 & 0.54 \\
&QMC & 0.28 & 0.54  \\
 &BOSampler & 0.25 & 0.45 \\
&BOSampler + QMC & 0.25 & 0.45   \\
\hline

\multirow{4}{*}{\textbf{PECNet}~\cite{pecnet}}& MC  & 0.32 & 0.56  \\
&QMC & 0.31 & 0.54  \\
 &BOSampler & 0.30 & 0.51  \\
&BOSampler + QMC &0.30 &0.50    \\
\hline
\multirow{4}{*}{\textbf{Social-STGCNN}~\cite{stgcnn}}& MC & 0.45 & 0.75  \\
&QMC & 0.39 & 0.65  \\
 &BOSampler & 0.41 & 0.69 \\
&BOSampler + QMC & 0.37 & 0.62  \\
\hline
\multirow{4}{*}{\textbf{STGAT}~\cite{huang2019stgat}}& MC & 0.46 & 0.90  \\
&QMC & 0.45 & 0.89  \\
 &BOSampler & 0.44 & 0.85 \\
&BOSampler + QMC  & 0.44& 0.84 \\
\hline\hline
\end{tabular}
\end{center}
\label{table:eth_ucy}
\vspace{-0.6cm}
\end{table}

\indent\textbf{Evaluation Metric.}
We follow the same evaluation metrics adopted by previous stochastic trajectory prediction methods~\cite{socialgan,pecnet,sgcn,huang2019stgat,mid}, which use widely-used evaluation metrics: minimal Average Displacement Error (minADE) and minimal Final Displacement Error (minFDE). ADE denotes the average error between all the ground truth positions and the estimated positions while FDE computes the displacement between the endpoints. Since the stochastic prediction model generates a finite set ($N$) of trajectories instead of the single one, we use the minimal ADE and FDE of $N = 20$ trajectories following~\cite{socialgan,trajectron++}, called Best-of-20 strategy. For the ETH-UCY dataset, we use the leave-one-out cross-validation evaluation strategy where four scenes are used for training and the remaining one is used for testing.
Besides, for all experiments, we evaluate methods 10 times and report the average performance for robust evaluation.

\indent\textbf{Baseline Methods.}
We evaluate our \ourmeos with five mainstream stochastic pedestrian trajectory prediction methods, including Social-GAN~\cite{socialgan}, PECNet~\cite{pecnet}, Trajectron++~\cite{trajectron++}, Social-STGCNN~\cite{stgcnn} and STGAT~\cite{huang2019stgat}. 
Social-GAN~\cite{socialgan} learns a GAN model with a normal Gaussian noise input to represent human multi-modality. \ourmeos optimizes the sampled noise to encourage diversity. 
STGAT~\cite{huang2019stgat} is an improved version of Social-GAN, which also learns a GAN model for motion multi-modality and applies the graph attention mechanism to encode spatial interactions.
PECNet~\cite{pecnet} applies the different endpoints to generate multiple trajectories. We optimize these end-points whose prior is the learned Endpoint VAE.
Trajectron++~\cite{trajectron++} uses the observation as the condition to learn a CVAE with  the learned discrete latent variable.
Social-STGCNN~\cite{stgcnn} directly learns parameters of the Gaussian distribution of each point and samples from it. Here, we can directly optimize the position of points.
All these baseline methods use the Monte Carlo (MC) sampling methods for generations. We can directly change the sampling manner from MC to our \ourmeos with their trained models, i.e. our method doesn't need any training data to refine the sampling process.
Beyond MC sampling, we also compare \ourmeos with Quasi-Monte Carlo (QMC) sampling introduced in~\cite{npsn}, which uses low-discrepancy quasi-random sequences to replace the random sampling.
It can generate evenly distributed points and achieve more uniform sampling.

\begin{figure*}
    \centering
    \includegraphics[width=\linewidth]{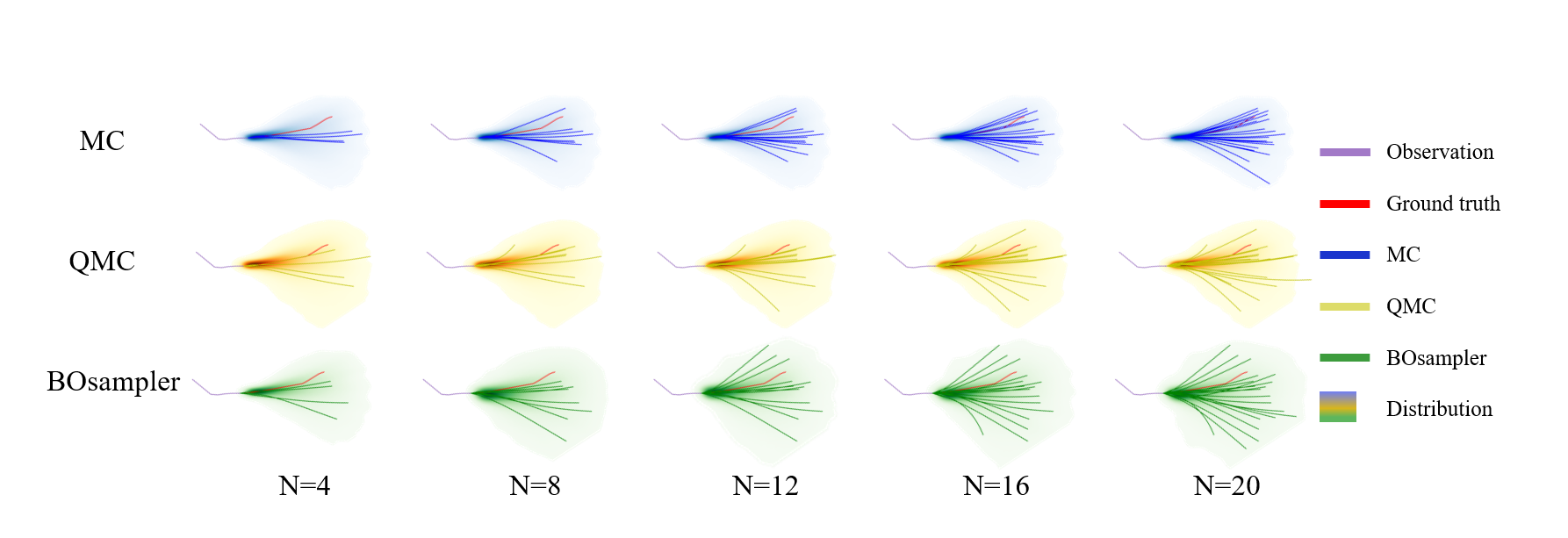}
    \caption{Visualization of trajectories with sample number $N = 4,8,12,16,20$, generated by MC, QMC and \ourmeos. Different from MC and QMC whose sampling distribution is the same with all sample numbers, \ourmeos adaptively modifies the sampling distribution with existing samples.}
    \label{fig: gradually}
    \vspace{-0.3cm}
\end{figure*}

\subsection{Quantitative Comparison}

\indent\textbf{Performance on the exception subset of ETH-UCY.}
The goal of our method is to help models to generate more comprehensive and reliable samples. Thus, we focus on abnormal situations such as turning left, turning right, or U-turn. Though these situations are the minority of all trajectories, they are still crucial for the applications such as intelligent transportation and auto-driving due to their safety and reliability.
The detailed selection procedure of the exception subset is explained in \cref{sec:exception_dataset}. As shown in \cref{table:exception}, we give minADE and minFDE results using the same pre-trained model across different sampling methods including MC, QMC, and \ourmeos, based on five baseline methods.
\ourmeos shows a significant improvement in exception trajectories compared to MC and QMC. The average performance gain rate of \ourmeos to MC on ADE/FDE among five baseline models is 23.71\% and 27.49\%, respectively. It implies that the promotion of \ourmeos over the original fixed pre-trained model mainly lies in the rare trajectories.

\indent\textbf{Performance on ETH-UCY.}
Beyond the exceptional cases, we also quantitatively compare \ourmeos with MC and QMC sampling methods on the original ETH-UCY dataset. 
As shown in \cref{table:eth_ucy}, we provide the minADE and minFDE results using the same pre-trained model and different sampling methods. 
Here, we only report the average results on all five scenes. Please kindly refer to the \textbf{supplementary materials} for the complete experimental results on each scene. 
For all baseline methods, \ourmeos consistently outperforms the MC sampling method, which shows the effectiveness of the proposed method, though not much. 
It is reasonable that all results from a fixed model with different sampling methods are comparable because only a small part of trajectories are uncommon (lie in low probability), while most testing trajectories are normal. But we want to highlight that these low-probability trajectories may raise safety risks for autonomous driving systems.
The results show that \ourmeos can provide a better prediction for possible low-probability situations without reducing the accuracy of most normal trajectories.
In addition, \ourmeos also shows an improvement over the QMC method on most baselines. For Social-STGCNN~\cite{stgcnn}, though \ourmeos achieves improvement over the MC method by a more considerable margin, it is still slightly lower than the QMC method. It is because Social-STGCNN adds the indeterminacy on each position, whose variable dimension is too large ($2\times 12 =24$) for Bayesian Optimization.
Furthermore, we also show that the proposed \ourmeos is not contradictory to the QMC method. Using QMC in the warm-up stage, we can further improve the performance of \ourmeos.
For example, for Social-STGCNN, \ourmeos + QMC can further improve the QMC method and achieve 0.37 ADE and 0.62 FDE.
Please note that we don't compare with the NPSN method~\cite{npsn} since it is a supervised method that needs to access the source data and re-train the models.



\subsection{Qualitative Comparison}
We further investigate our method with three qualitative experiments. Firstly, we visualize the sampled trajectories of MC, QMC, and \ourmeos with different sample numbers. Secondly, we visualize and compare the best predictions among sampled trajectories of MC, QMC, and \ourmeos in the real environment.
Thirdly, we also provide the visualization of some failure cases.


\indent\textbf{Trajectories with different sample numbers.}
In this experiment, we aim to investigate how the sampling (posterior) distribution changes with the increase of the sampling number.
As shown in Figure~\ref{fig: gradually}, we provide the sampling results of MC, QMC, and \ourmeos with sample number $N =4,8,12,16,20$, where the light area denotes the sampling (posterior) distribution.
We can observe that the sampling distributions of both MC and QMC are unchanged. The only difference is that QMC smooths the original distribution. It indicates that QMC may not work well when the sampling number is small since the distribution is changed suddenly. 
Unlike them, \ourmeos gradually explore the samples with low probability with the increase of the sample number, which can achieve an adaptive balance between diversity and accuracy. 
When the sampling number $N$ is small, \ourmeos tends to sample close to the prior distribution. When $N$ is larger, the model is encouraged to select those low-probability samples.


\begin{figure}
    \centering
    \includegraphics[width=\linewidth]{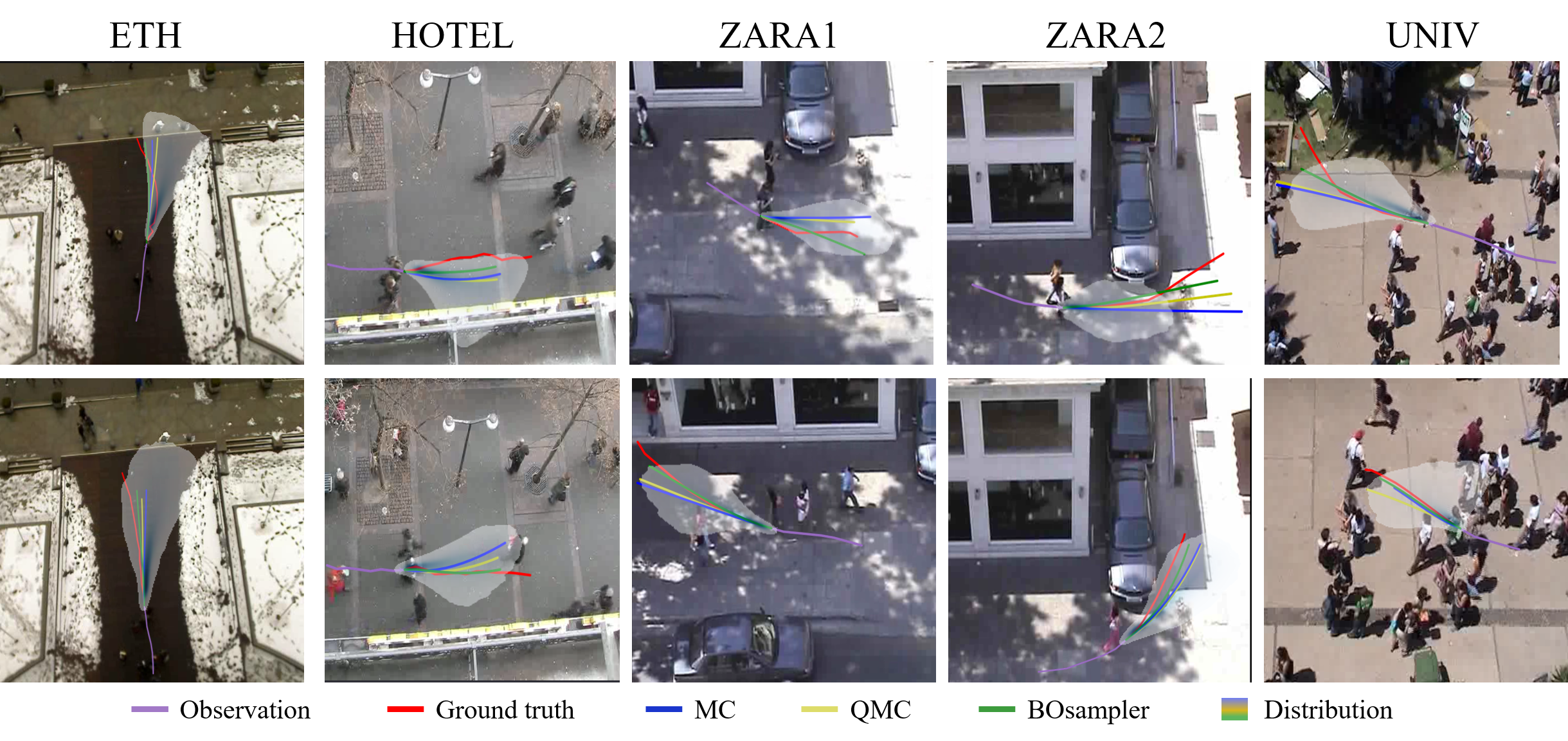}
    \caption{Visualization of our method in five datasets. We sampled 20 times with MC, QMC, and BOsampler and compared the best-predicted trajectories from the sampled results. And the light areas are density graphed generated by sampling 2000 times with MC.}
    \label{fig: best}
    \vspace{-0.5cm}
\end{figure}

\begin{figure}[t]
    \centering
    \includegraphics[width=\linewidth,]{ 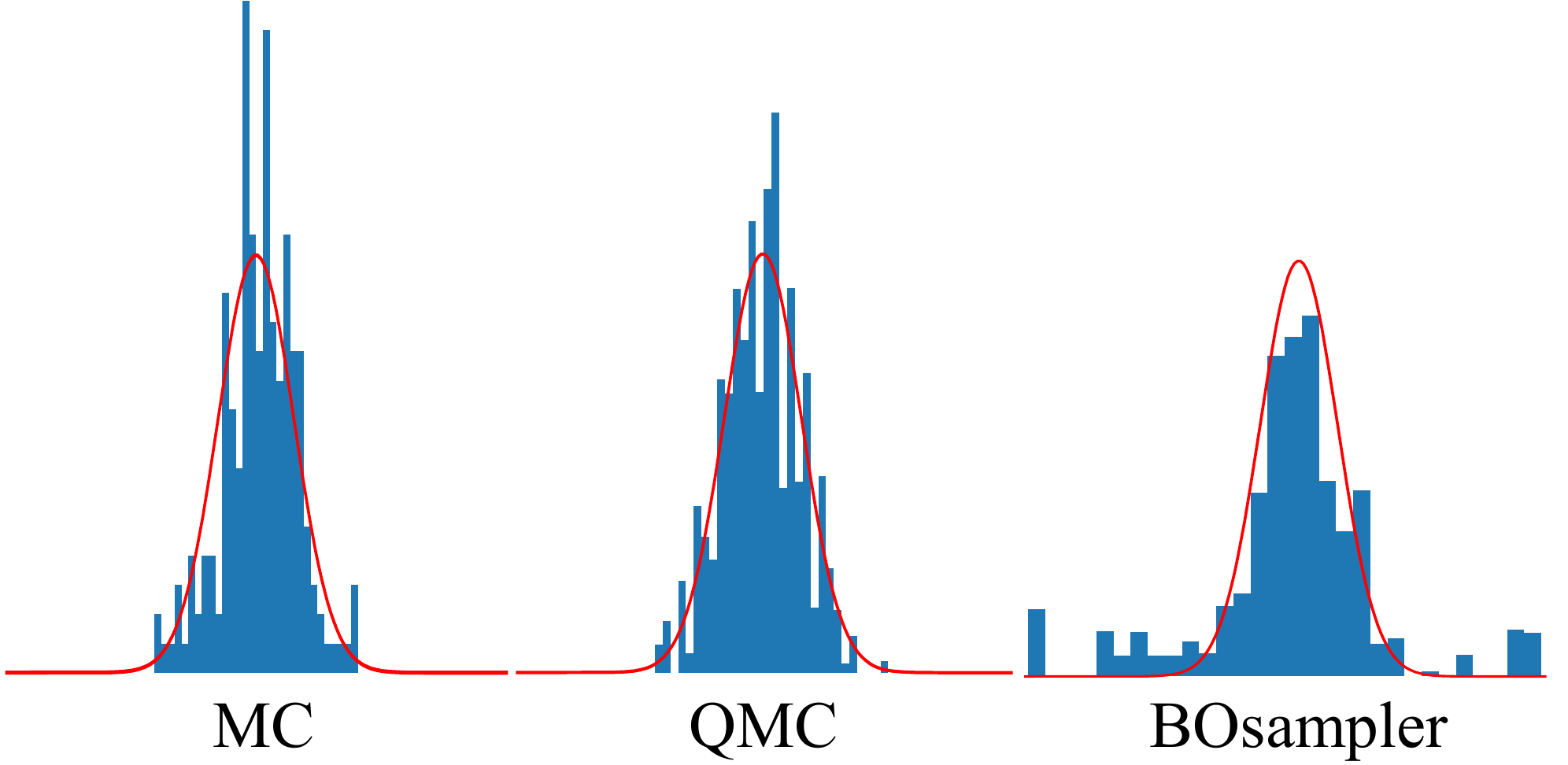}
    \caption{Frequency histogram of MC, QMC, and BOsampler}
    \label{fig: comparison}
    \vspace{-0.3cm}
\end{figure}

\indent\textbf{Visualization.}
We also compare the best predictions of different sampling methods to provide an intuition in which situation \ourmeos works well. As shown in Figure~\ref{fig: best}, we give the best predictions of different sampling methods and the ground truth trajectory on five scenes in the ETH-UCY dataset. We observe that \ourmeos can provide the socially-acceptable paths in the low-probabilities (away from normal ones). For example, when the pedestrian turns left or right, the gourd truth will be far away from the sampled results of MC and QMC, but our method's sampled results are usually able to cover this case.
It indicates that \ourmeos encourage the model to explore the low-probability choices. Besides, we also provide the visualization of the failure cases to understand the method better. We found that \ourmeos may lose the ground truth trajectory when the most-likely prediction is far away from the ground truth.

Besides, as shown in Figure~\ref{fig: comparison}, we visualize the optimized sampling distributions of MC, QMC, and \textbf{\texttt{BOsampler}} with the original standard Gaussian distribution $ \mathcal{N}(0,1)$. By the simulation results, we show that BOsampler can mitigate the long-tail property, while MC and QMC cannot.

\begin{figure}[t]
    \centering
    {
    \includegraphics[width= 0.45\linewidth,trim=0 5 0 0,clip]{ 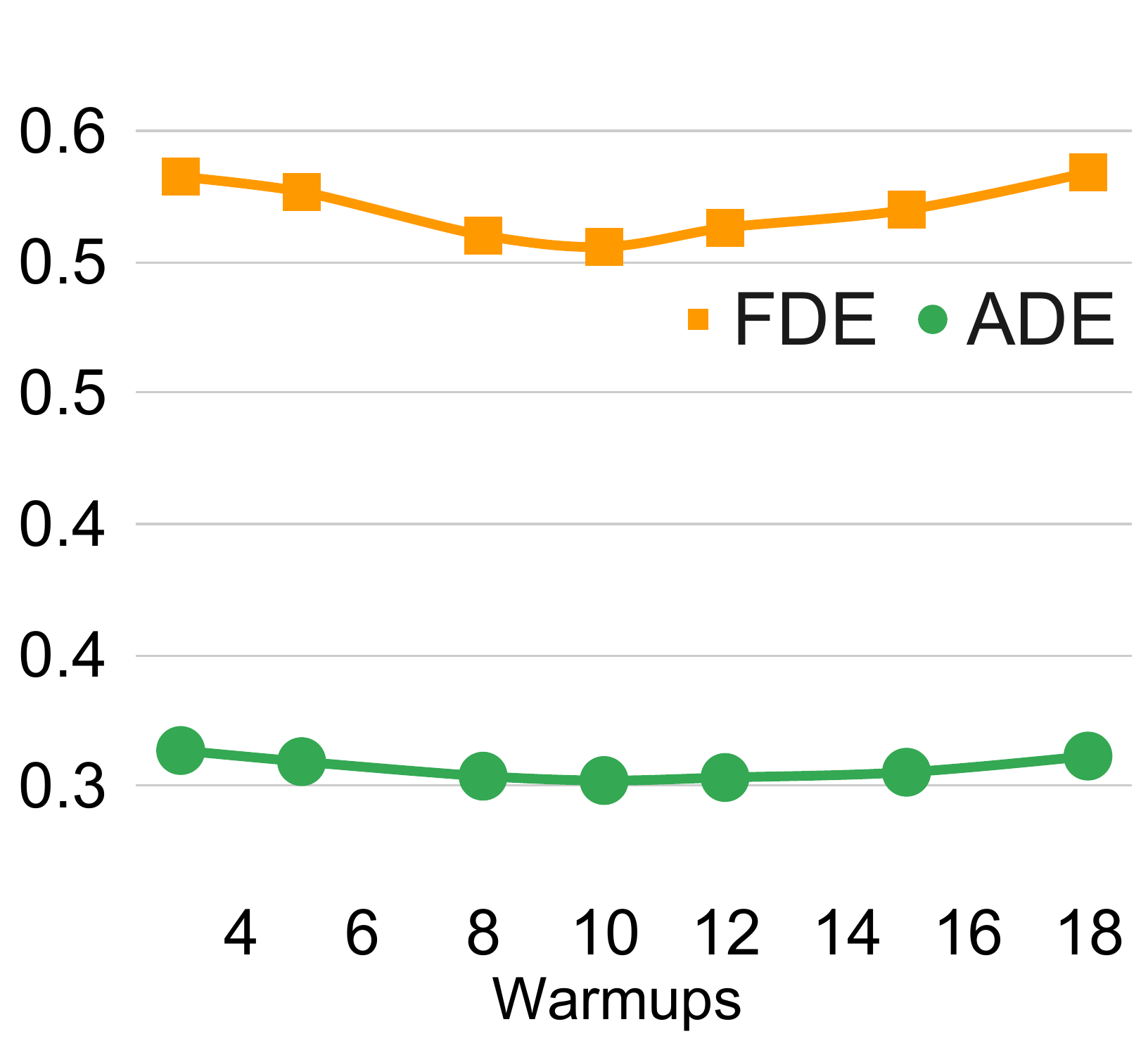} \quad \quad
    \includegraphics[width = 0.45\linewidth,trim=0 5 0 0,clip]{ 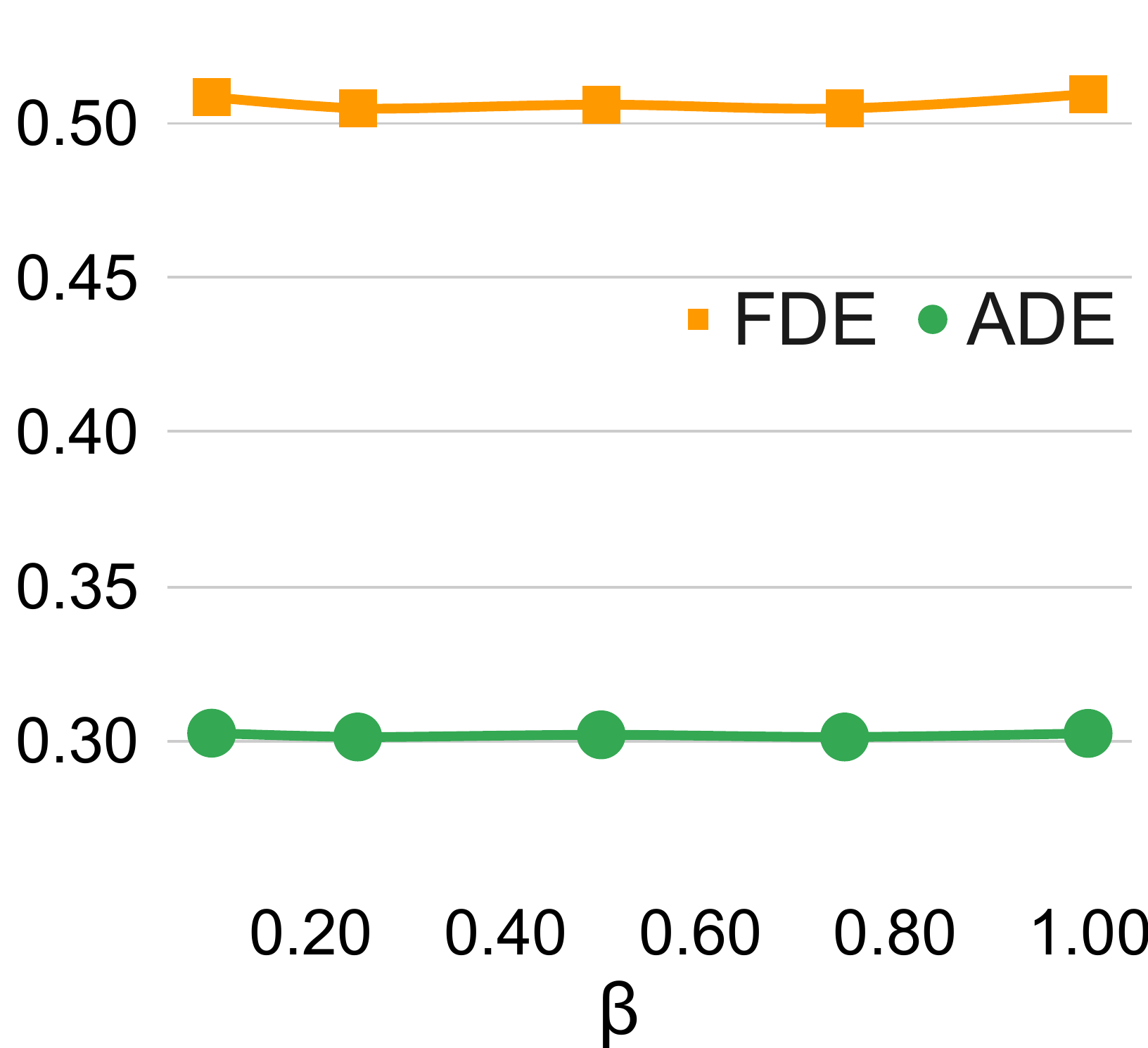} 
    }
    \vspace{-0.2cm}
    \caption{Left: ADE and FDE with different warm-up samples by \ourmeos on PECNet. \\ Right: ADE and FDE with different hyper-parameters $\beta$ by \ourmeos on PECNet. }
    \label{fig:4_4_1}
    \vspace{-0.3cm}
\end{figure}

\subsection{Ablation Studies and Parameters Analysis}
In this subsection, we conduct ablation studies and parameters analysis to investigate the robustness of different hyper-parameters.
Then, we provided a detailed analysis of the sampling process with a different number of samples.

\indent\textbf{Analysis with warm-up:} \label{sec: warmup}
We choose among the number of warm-up $\text{w} = 3, 5, 8, 10, 12, 15, 18$, and then we use PECNet as the baseline model. The results on the ETH/UCY dataset with the Best-of-20 strategy are shown in the top row of Figure~\ref{fig:4_4_1}. For all the numbers of warm-ups, \ourmeos achieves better performance than the MC baseline. When the number of warm-ups is close to half of the number of samples, which is 10, the corresponding ADE/FDE is better than other options. Although decreasing the number of warm-ups will encourage more exploration by increasing the number of \ourmeos, the performance overall will be hurt because the abnormal trajectories only make up a relatively small portion of the entire dataset. Setting the number of warm-up to half of the number of entire samples helps balance exploration and exploitation.

\indent\textbf{Analysis with acquisition function:} 
We analyze the robustness of the hyper-parameter by selecting $\beta \in [0.1,1]$, separated evenly in this range. We choose PECNet as a base model and use the Best-of-20 strategy to evaluate on the ETH/UCY dataset. The performance is close among five acquisition factors $\beta$, which means the performance of \ourmeos is stable when the acquisition factor is set within a reasonable range.

\begin{figure}
    \centering
    \subcaptionbox{Social GAN\label{subfig:sgan_num_of_samples}}{
    \includegraphics[width= 0.33\linewidth]{ 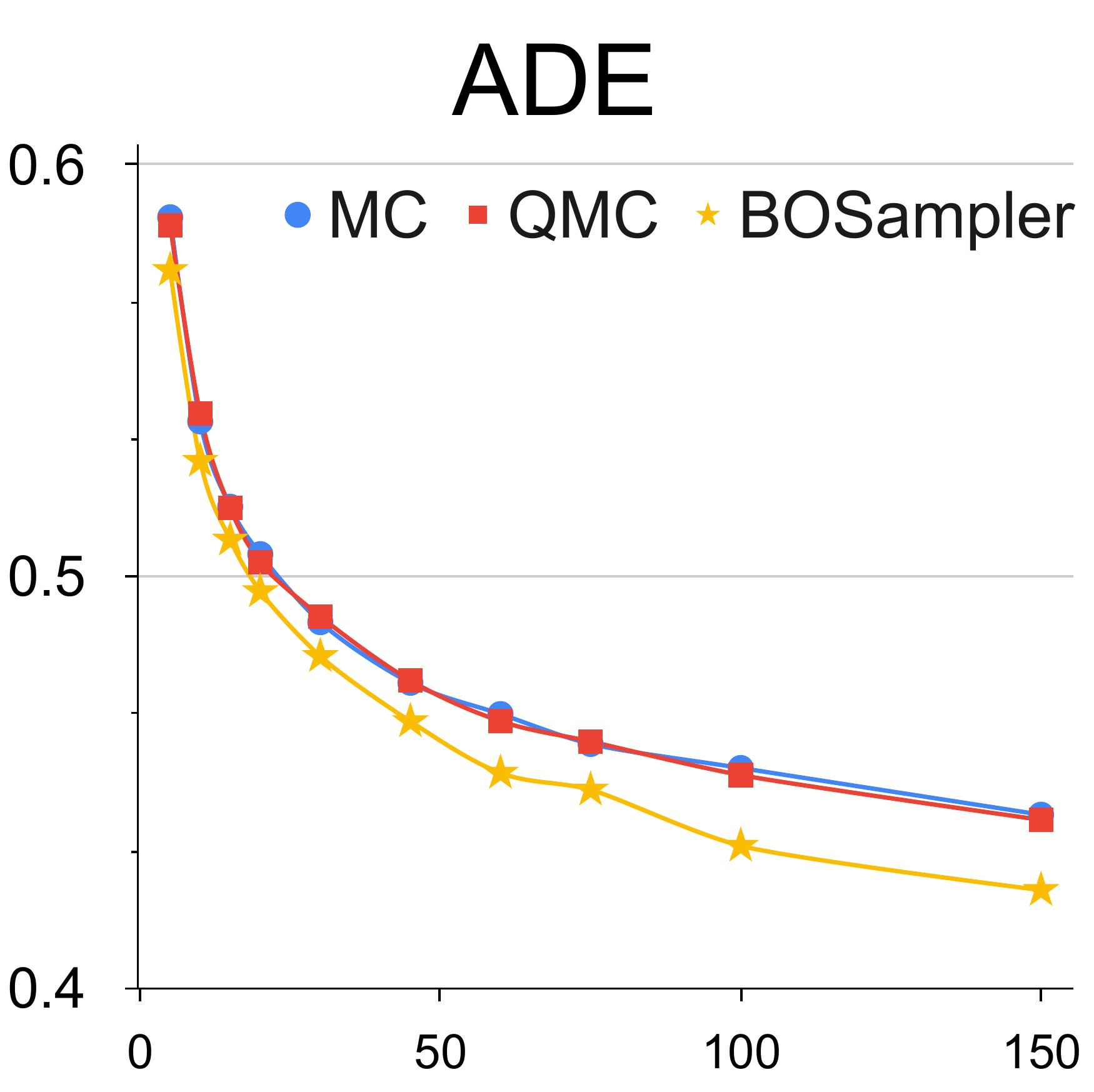} 
    \includegraphics[width = 0.33\linewidth]{ 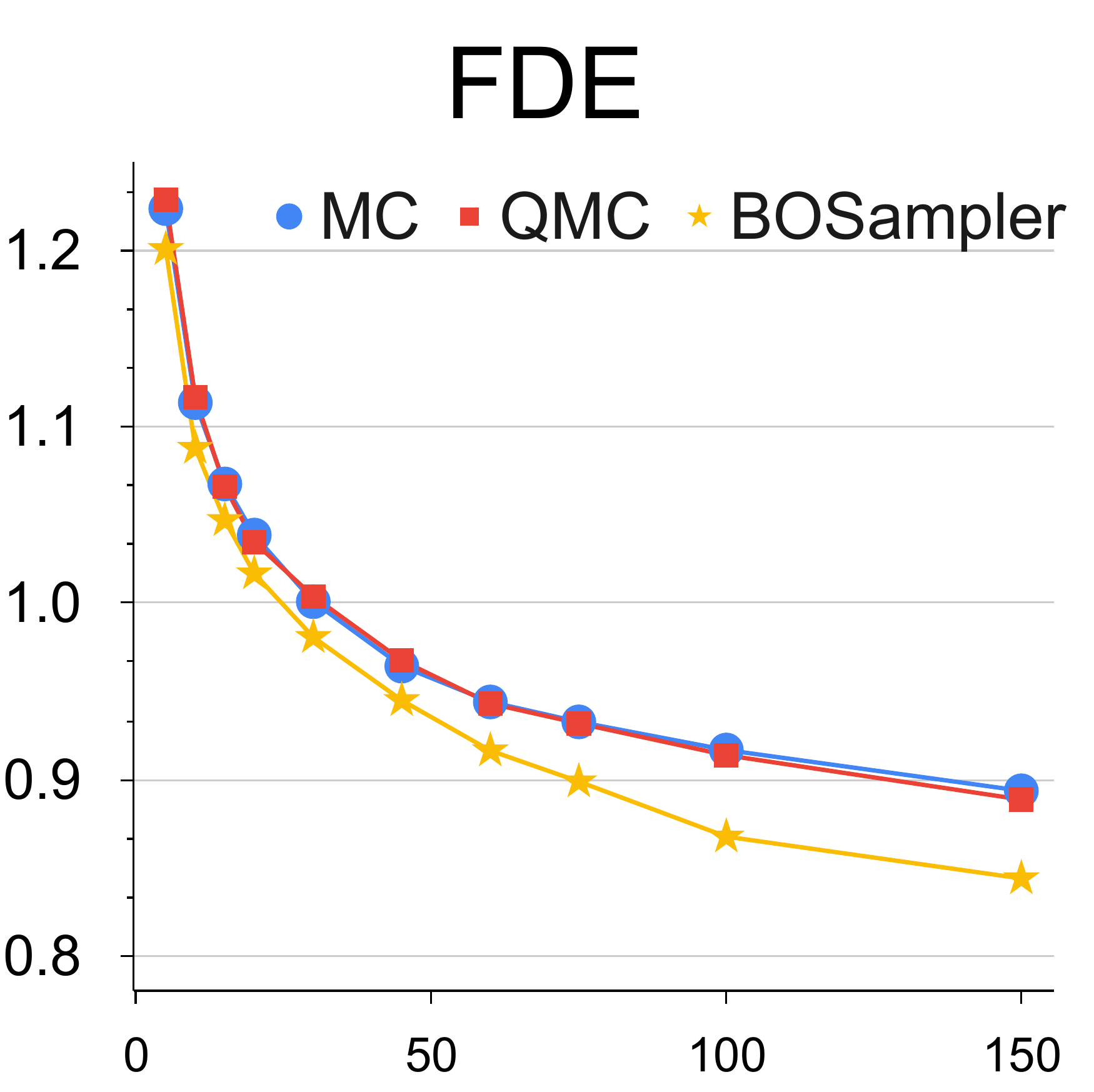} 
    \includegraphics[width = 0.33\linewidth]{ 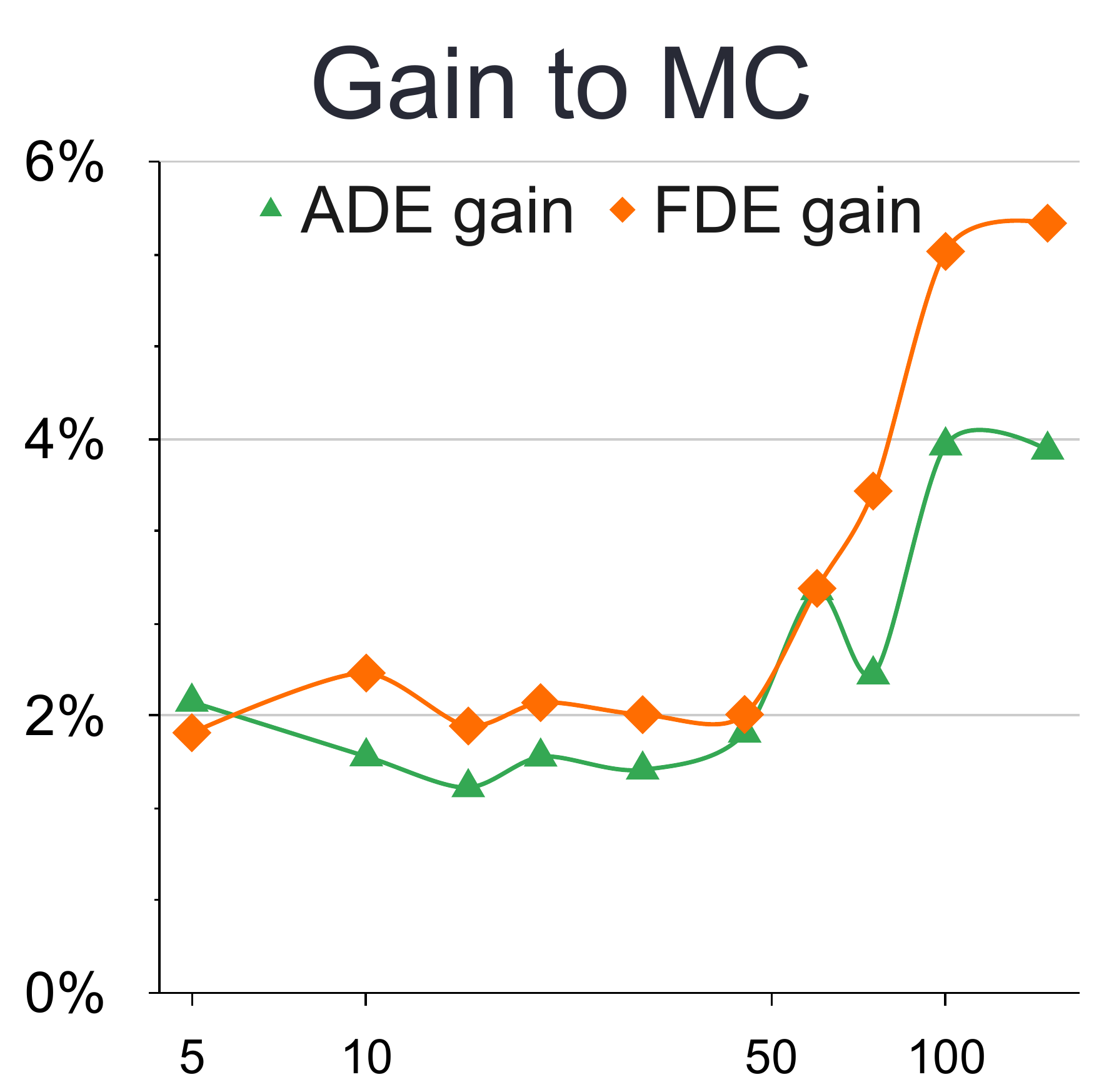}
    }
    \subcaptionbox{PECNet\label{subfig:pecnet_num_of_samples}}{
    \includegraphics[width= 0.33\linewidth]{ 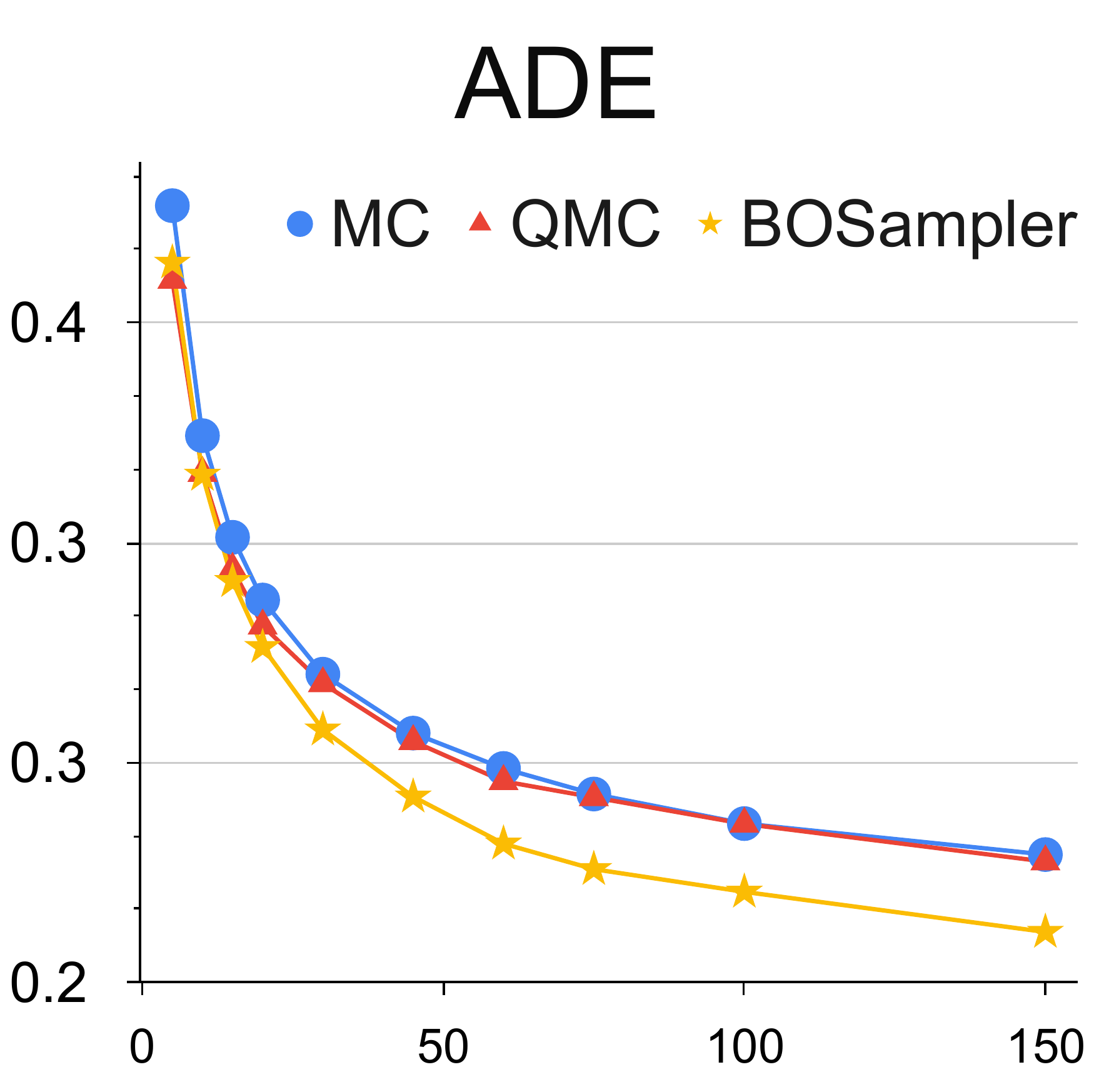} 
    \includegraphics[width = 0.33\linewidth]{ 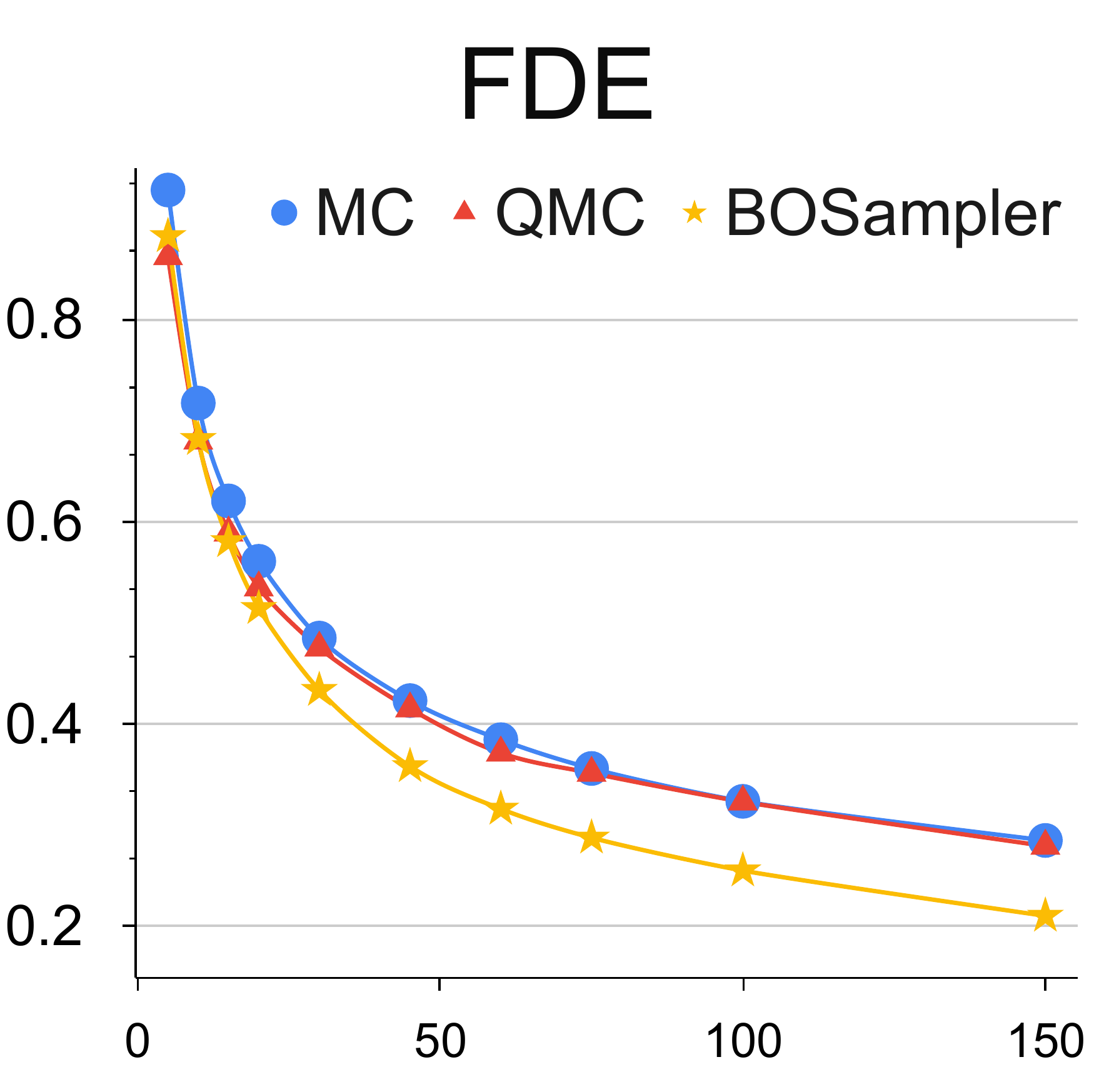} 
    \includegraphics[width = 0.33\linewidth]{ 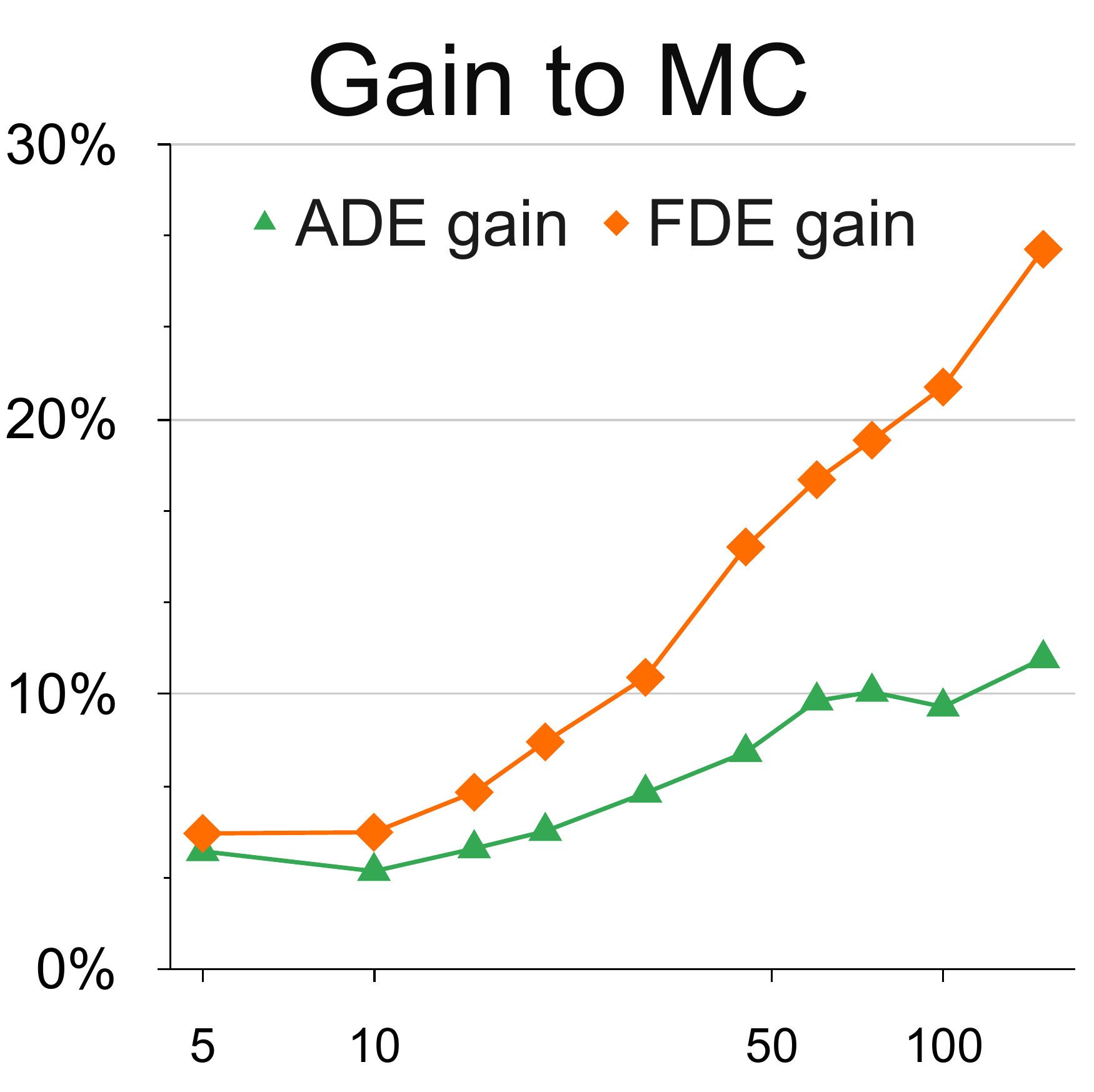}
    }
    \caption{ADE, FDE, and performance gain of \ourmeos to MC on Social GAN and PECNet across a different number of samples.}
    \label{fig:4_4_num_samples}
    \vspace{-0.3cm}
\end{figure}

\indent\textbf{Analysis with different number of samples:}
We provide this quantitative experiment with respect to the number of samples to better understand the simpling process of our \ourmeos. As shown in Figure~\ref{fig:4_4_num_samples}, we compare the ADE and FDE on the ETH-UCY dataset of MC, QMC, and \ourmeos with different numbers of samples on Social GAN and PECNet. We choose a number of samples $N=5,10,15,20,30,45, 60,75,100,150$. \ourmeos works well in all settings, which demonstrates an adaptive balance between diversity and accuracy. It also shows that \ourmeos will work even if the warm-up steps are extremely small (less than five as is shown in this case). Besides, we find that our \ourmeos obtains a larger improvement over MC than the improvement of QMC when the number of samples increases.
With the Gaussian process, \ourmeos can gradually refine the posterior distribution with the sampling process.

\section{Conclusion}
In this paper, we have proposed an unsupervised sampling method, called {\textbf{\texttt{BOsampler}}}, to promote the sampling process of the stochastic trajectory prediction system. In this method, we formulate the sampling as a sequential Gaussian process, where the current prediction is conditioned on previous samples. Using Bayesian optimization, we defined an acquisition function to explore potential paths with low probability adaptively. Experimental results demonstrate the superiority of \ourmeos over other sampling methods such as MC and QMC.

\textbf{Broader Impact \& limitations:} \ourmeos can be integrated with existing stochastic trajectory prediction models without retraining.
It provides reasonable and diverse trajectory sampling, which can help the safety and reliability of intelligent transportation and autonomous driving. Despite being training-free, this inference time sampling promoting method still requires a time cost due to sequential modeling. Taking Social GAN as a baseline, our method needs $8.56s$ for predicting 512 trajectories while MC needs $4.92s$. 
Better computational techniques may mitigate this issue.

\section*{Acknowledgments}
This project was partially supported by the National Institutes of Health (NIH) under Contract R01HL159805, by the NSF-Convergence Accelerator Track-D award \#2134901, by a grant from Apple Inc., a grant from KDDI Research Inc, and generous gifts from Salesforce Inc., Microsoft Research, and Amazon Research. We would like to thank our colleague Zunhao Zhang from MBZUAI for providing computation resource for part of the experiments.

{\small
\bibliographystyle{ieee_fullname}
\bibliography{bosampler}
}

\clearpage
\appendix

\noindent \textit{\large \bf Appendix for BoSampler}

\vspace{.1cm}

\section{The algorithm of \ourmeos}
We give the algorithm to give a short but clear summarization of \ourmeos. The detailed algorithm is shown in \cref{alg:cap}.

\begin{algorithm}[h]
\caption{Sampling procedure of \ourmeos}\label{alg:cap}
\begin{algorithmic}
\Require generator G, observed trajectory $X$, pseudo score evaluation function f.
\For{$n$ in \{1,...,w\}}            \Comment{Warm-up}
    \State Sample $\text{z}_n \sim p_{\text{z}}$
    \State $\hat{Y}_n \gets G(X,\text{z}_n)$
\EndFor

\For{$n$ in \{w+1,...,N\}}            \Comment{\ourmeos}
    \State Fit the Gaussian Process $\mathcal{GP}$ by $\text{z}_{1:n-1}$ and $f(\text{z}_{1:n-1})$ 
    \State Use the posterior of $\mathcal{GP}$ to build the acquisition function $\phi_n (\text{z})$
    \State $\text{z}_n \gets \underset{\text{z}}{\operatorname{argmax}} {\phi}_n (\text{z}) $
    \State $\hat{Y}_n \gets G(X,\text{z}_n)$
\EndFor
\State Return $\hat{Y}=\{\hat{Y}_{n} | n\in [1,...,N]\}$
\end{algorithmic}
\end{algorithm}
\vspace{-0.1cm}

\section{Complete quantitative experimental results on ETH/UCY}
We present the complete experiment results on five scenes of ETH/UCY datasets in \cref{table:eth_ucy_full}. For all baseline methods, \ourmeos consistently outperforms the MC sampling method, which shows the effectiveness of the proposed method, though not much. \ourmeos also shows an improvement over the QMC method on most baselines. \ourmeos doesn't have the same level of performance gain in the complete set of ETH/UCY compared to the exception subset. It is because the uncommon trajectories only comprise a small portion of the dataset. \ourmeos achieves more performance gain on the ETH dataset. The reason is probably the same: uncommon trajectories show up more frequently in ETH dataset than in the other four datasets in ETH/UCY. Although \ourmeos does not achieve a huge improvement in the complete set, considering \ourmeos has a significant gain in the exception subset, it is clear that \ourmeos balances exploitation of the baseline model's distribution and exploration of the edge cases. So \ourmeos helps promote the robustness of the sampling process of baseline models. We highlight that \ourmeos focus on trajectories with low probability. Although these cases are the minority, it is meaningful and crucial to consider them in intelligent transportation and autonomous driving. 

\begin{figure}[t]
    \centering
    \includegraphics[width=8cm]{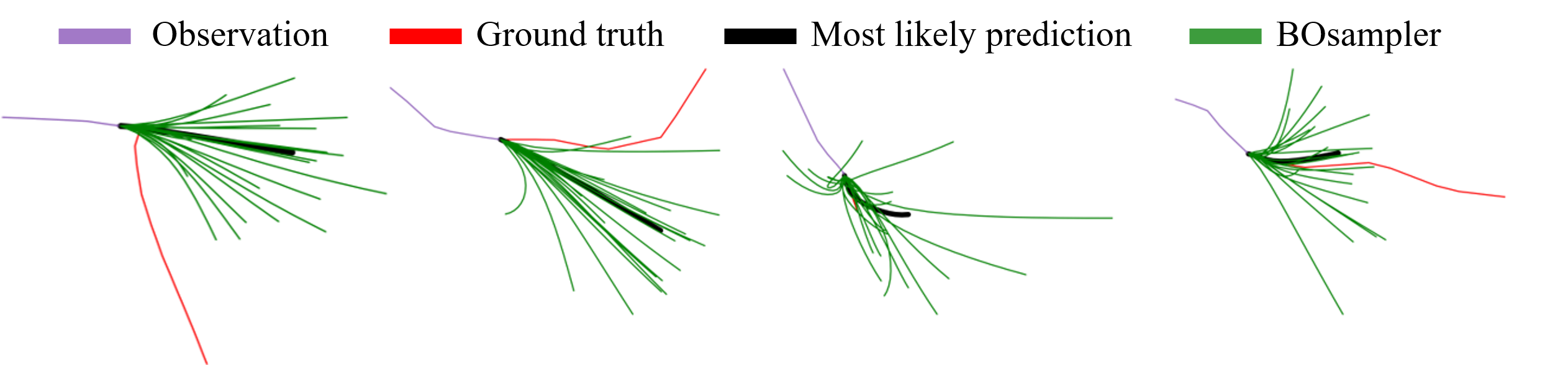}
    \caption{Failure cases: In the first and second graph, we show that when a pedestrian changes direction suddenly, our method may not predict such a change. In the third graph, when a pedestrian stops suddenly (e.g. a pram gets stuck in the ground), or when a pedestrian suddenly starts walking, our method may also fail. }
    \label{fig:failure_case}
    \vspace{-0.3cm}
\end{figure}

\begin{table}[h]
\centering
\vspace{-0.0cm}

\vspace{-0.0cm}
\renewcommand\arraystretch{1.0}
\renewcommand\tabcolsep{1.5pt}
\begin{tabular}{c|cccccc}
\hline
 & \multicolumn{2}{c}{top 4\%} & \multicolumn{2}{c}{top 12\%}  & \multicolumn{2}{c}{full} \\ \cline{2-3} \cline{4-5}\cline{6-7}
 \hline
 MC &   1.21/2.33 &   & 0.88/1.76    &    & 0.32/0.56 &     \\ 
 QMC &  1.20/2.33 & &   0.86/1.70    &       & 0.31/0.54 &  \\ 
 \ourmeos & 0.97/1.75  &  &     0.74/1.38   &     & 0.30/0.51  &          \\
\hline

\end{tabular}

\vspace{-0.0cm}
\caption{The results of PECNet on ETH-UCY with different exception ratios.}
\label{table: pec}
\end{table}

\section{ Experiments on exception subsets with different ratios}
In Section 4.2 of the main manuscript, we set the ratio of the exception set as 4\%. In order to verify the validity of the chosen subset, we further provide a 12\% variation, which demonstrates a consistent performance trend with the original subset, thus confirming the representativeness and appropriateness of the ratio setting of our experiments. The experiment results are presented in \cref{table: pec}. When the ratio goes to 12\%, \ourmeos still achieves considerably better performance than baselines.

\section{Visualization of the failure cases}

We provide the visualization of the failure cases to better understand the method.  From the visualization in \cref{fig:failure_case}, we found \ourmeos may lose the ground truth trajectory when the most-likely prediction is far away from the ground truth. It is not surprising since our method is based on the assumption of the good prior distribution. Solving this problem without accessing more data is not trivial. A potential solution is to modify the prior distribution using the testing data during the inference~\cite{ttt}.

\begin{table*}[t]
\caption{Quantitative results on the ETH/UCY dataset with Best-of-20 strategy in ADE/FDE metric. Lower is better. * updated version of \href{https://github.com/StanfordASL/Trajectron-plus-plus/issues/26}{Trajectron++ }}
\vspace{-0.2cm}
\linespread{3.0}
\renewcommand\arraystretch{1.1}
\renewcommand\tabcolsep{3pt}
\begin{center}
\newcolumntype{g}{>{\columncolor{Gray}}c}
\newcolumntype{y}{>{\columncolor{LightCyan}}c}
\newcolumntype{d}{>{\columncolor{DarkCyan}}c}
\begin{tabular}{l| c|c g c g c g c g c g y d}
\hline
\hline
\multirow{2}{*}{\textbf{Baseline Model}} &\multirow{2}{*}{\textbf{Sampling}} &\multicolumn{2}{c}{\textbf{ETH}} & \multicolumn{2}{c}{\textbf{HOTEL}} &  \multicolumn{2}{c}{\textbf{UNIV}} &  \multicolumn{2}{c}{\textbf{ZARA1}} & \multicolumn{2}{c}{\textbf{ZARA2}} & \multicolumn{2}{c}{\textbf{AVG}} \\ 
\cline{3-14}
~ & &\multicolumn{1}{c}{\textbf{ADE}} & \multicolumn{1}{c}{\textbf{FDE}} &  \multicolumn{1}{c}{\textbf{ADE}} &  \multicolumn{1}{c}{\textbf{FDE}} & 
\multicolumn{1}{c}{\textbf{ADE}} &  \multicolumn{1}{c}{\textbf{FDE}} &
\multicolumn{1}{c}{\textbf{ADE}} &  \multicolumn{1}{c}{\textbf{FDE}} &
\multicolumn{1}{c}{\textbf{ADE}} &  \multicolumn{1}{c}{\textbf{FDE}} &
\multicolumn{1}{c}{\textbf{ADE}} &  \multicolumn{1}{c}{\textbf{FDE}} 

\\
\hline
\multirow{4}{*}{\textbf{Social-GAN}~\cite{socialgan}}& MC &0.77&1.40 & 0.43&0.88 & 0.75& 1.50  &0.35&0.69   & 0.36& 0.72 & 0.53&1.05\\
&QMC & 0.76 & 1.38 & 0.43 & 0.87 & 0.75& 1.50 & 0.34 & 0.69& 0.35 & 0.72  & 0.53 & 1.03 \\
 &BOSampler & 0.73 & 1.28 & 0.43 & 0.87 &0.75 & 1.50& 0.34& 0.69& 0.35& 0.71& 0.52 & 1.01 \\
&BOSampler + QMC  & 0.72 & 1.26 & 0.43 & 0.87 &0.74 & 1.49 & 0.34& 0.69 & 0.35 & 0.71 & 0.52 & 1.00   \\
\hline
\multirow{4}{*}{\textbf{Trajectron++}~\cite{trajectron++}}& MC & 0.43 & 0.86 & 0.12 & 0.19 & 0.22 & 0.43 & 0.17 & 0.32 & 0.12 & 0.25 & 0.21 & 0.41  \\
&QMC & 0.43 & 0.84 & 0.12 & 0.19 & 0.22 & 0.42 & 0.17 & 0.31 & 0.12 & 0.24 & 0.21 & 0.40  \\
 &BOSampler & 0.34 & 0.64 & 0.12 & 0.21 & 0.18 & 0.40 & 0.14 & 0.30 & 0.11 & 0.24 & 0.18 & 0.36  \\
&BOSampler + QMC & 0.34 & 0.64 & 0.12 & 0.21 & 0.18 & 0.40 & 0.14 & 0.30 & 0.11 & 0.24 & 0.18 & 0.36   \\
\hline
\multirow{4}{*}{\textbf{Trajectron++}~\cite{trajectron++}\textsuperscript{*}}& MC & 0.57 & 1.06 & 0.16 & 0.26 & 0.30 & 0.61 & 0.22 & 0.42 & 0.16 & 0.33 & 0.28 & 0.54  \\
&QMC &  0.57 & 1.05 & 0.16 & 0.26 & 0.30 & 0.61 & 0.22 & 0.42 & 0.16 & 0.33 & 0.28 & 0.54  \\
 &BOSampler & 0.49 & 0.82 & 0.15 & 0.23 & 0.27 & 0.53 & 0.20 & 0.38 & 0.15 & 0.29 & 0.25 & 0.45 \\
&BOSampler + QMC & 0.49 & 0.82 & 0.15 & 0.23 & 0.27 & 0.53 & 0.20 & 0.38 & 0.15 & 0.29 & 0.25 & 0.45 \\
\hline
\multirow{4}{*}{\textbf{PECNet}~\cite{pecnet}}& MC & 0.61 & 1.07 & 0.22 & 0.39 & 0.34 & 0.56 & 0.25 & 0.45 & 0.19 & 0.33 & 0.32 & 0.56  \\
&QMC & 0.60 & 1.04 & 0.21 & 0.38 & 0.33 & 0.53 & 0.24 & 0.43 & 0.18 & 0.31 & 0.31 & 0.54  \\
 &BOSampler & 0.56 & 0.92 & 0.21 & 0.38  & 0.32 & 0.52 & 0.24 & 0.42 & 0.18 & 0.31 & 0.30 & 0.51  \\
&BOSampler + QMC & 0.56 & 0.91 & 0.21 & 0.37 & 0.31 & 0.51 & 0.24 &0.41 &0.18 & 0.31 &0.30 &0.50    \\
\hline
\multirow{4}{*}{\textbf{Social-STGCNN}~\cite{stgcnn}}& MC & 0.65 & 1.10 & 0.50 & 0.86 & 0.44 & 0.80 & 0.34 & 0.53 & 0.31 & 0.48 & 0.45 & 0.75  \\
&QMC & 0.62 & 1.03 & 0.38 & 0.57 & 0.36 & 0.63 & 0.32 & 0.52 & 0.29 & 0.50 & 0.39 & 0.65  \\
 &BOSampler & 0.57 & 0.90 &  0.44& 0.82& 0.43 & 0.76 & 0.34 & 0.54 & 0.26 & 0.45 & 0.41 & 0.69 \\
&BOSampler + QMC &0.49 & 0.74 & 0.39 & 0.73 & 0.41 & 0.72 & 0.32 & 0.52 & 0.26 & 0.40 & 0.37 & 0.62  \\
\hline
\multirow{4}{*}{\textbf{STGAT}~\cite{huang2019stgat}}& MC & 0.74 & 1.34 & 0.35 & 0.68 & 0.56 & 1.20 & 0.34 & 0.68 & 0.29 & 0.59 & 0.46 & 0.90  \\
&QMC & 0.73 & 1.32 & 0.35 & 0.67 & 0.56 & 1.20 & 0.34 & 0.68 & 0.29 & 0.59 & 0.45 & 0.89  \\
 &BOSampler & 0.70 & 1.15 &  0.35 & 0.67 & 0.55 & 1.17 & 0.34 & 0.68 & 0.29 & 0.59 & 0.44 & 0.85 \\
&BOSampler + QMC & 0.68 &	1.11 &	0.35 &	0.67 &	0.55&	1.17 &	0.33&	0.67 &	0.30 &	0.59 & 0.44	& 0.84 \\
\hline\hline
\end{tabular}
\end{center}
\label{table:eth_ucy_full}
\vspace{-0.3cm}
\end{table*}





\end{document}